\documentclass[10pt,twocolumn,letterpaper]{article}

\usepackage{cvpr}              

\usepackage{graphicx}
\usepackage{amsmath}
\usepackage{amssymb}
\usepackage{booktabs}
\usepackage{enumitem}
\usepackage{amsthm}
\usepackage{multirow}

\usepackage{comment}
\usepackage{algorithm}
\usepackage{algpseudocode}
\usepackage{booktabs}

\usepackage{graphicx}
\usepackage{mathrsfs}
\usepackage{booktabs}
\usepackage{tabularx}
\usepackage{makecell,rotating}
\usepackage{wrapfig}
\usepackage{array}
\usepackage{bbding}
\usepackage{pifont}
\usepackage{bm}

\usepackage{xcolor}

\newcommand{\ours}{LDGM}
\newcommand{\oursfull}{Layout Diffusion Generative Model}

\newcommand{\bx}{\mathbf{x}}

\newcommand{\ieno}{\textit{i}.\textit{e}.}
\newcommand{\egno}{\textit{e}.\textit{g}.} 

\definecolor{ForestGreen}{RGB}{0, 179, 45}
\definecolor{myred}{RGB}{243, 45, 136}
\definecolor{Gray}{RGB}{147, 171, 247}

\definecolor{mygray}{gray}{0.4}

\usepackage[pagebackref,breaklinks,colorlinks]{hyperref}

\usepackage[capitalize]{cleveref}
\crefname{section}{Sec.}{Secs.}
\Crefname{section}{Section}{Sections}
\Crefname{table}{Table}{Tables}
\crefname{table}{Tab.}{Tabs.}

\def\cvprPaperID{5507} 
\def\confName{CVPR}
\def\confYear{2023}

\begin{document}

\title{Unifying Layout Generation with a Decoupled Diffusion Model}


\author{Mude Hui\textsuperscript{\rm 1}\thanks{This work was done when Mude Hui was an intern at MSRA.} \quad Zhizheng Zhang\textsuperscript{\rm 2}\quad Xiaoyi Zhang\textsuperscript{\rm 2}\quad Wenxuan Xie\textsuperscript{\rm 2}\quad Yuwang Wang\textsuperscript{\rm 3}\quad Yan Lu\textsuperscript{\rm 2}\\
\textsuperscript{\rm 1}{Xi’an Jiaotong University}\quad
\textsuperscript{\rm 2}{Microsoft Research Asia}\quad
\textsuperscript{\rm 3}{Tsinghua University}\\
\tt\small \{zhizzhang,\ xiaoyizhang,\ wenxie,\ yanlu\}@microsoft.com \\ 
\tt\small theflood@stu.xjtu.edu.cn \quad
\tt\small wang-yuwang@mail.tsinghua.edu.cn}


\maketitle

\begin{abstract}

Layout generation aims to synthesize realistic graphic scenes consisting of elements with different attributes including category, size, position, and between-element relation. It is a crucial task for reducing the burden on heavy-duty graphic design works for formatted scenes, e.g., publications, documents, and user interfaces (UIs). Diverse application scenarios impose a big challenge in unifying various layout generation subtasks, including conditional and unconditional generation. In this paper, we propose a Layout Diffusion Generative Model (LDGM) to achieve such unification with a single decoupled diffusion model. LDGM views a layout of arbitrary missing or coarse element attributes as an intermediate diffusion status from a completed layout. Since different attributes have their individual semantics and characteristics, we propose to decouple the diffusion processes for them to improve the diversity of training samples and learn the reverse process jointly to exploit global-scope contexts for facilitating generation. As a result, our LDGM can generate layouts either from scratch or conditional on arbitrary available attributes. Extensive qualitative and quantitative experiments demonstrate our proposed LDGM outperforms existing layout generation models in both functionality and performance.

\end{abstract}

~
\section{Introduction}
\label{sec:intro}
Layout determines the placements and sizes of primitive elements on a page of formatted scenes (\egno, publications, documents, 
UIs), which has critical impacts on how viewers understand and interact with the information in this page~\cite{kong2021blt}.
Layout generation is an emerging task of synthesizing 
realistic and attractive graphic scenes with primitive elements of different categories, sizes, positions, and relations. It is of high demands for reducing the burden on heavy-duty graphic design works in diverse application scenarios.
Recently, there have been some research works studying unconditional generation~\cite{li2019layoutgan,arroyo2021variational,gupta2021layouttransformer,yamaguchi2021canvasvae,jiang2022coarse}, conditional generation based on user specified inputs (\egno, element types~\cite{lee2020neural,kong2021blt,kikuchi2021constrained}, element types and sizes~\cite{li2019layoutgan,kong2021blt} or element relations~\cite{lee2020neural,kikuchi2021constrained}),
conditional refinement based on coarse attributes~\cite{rahman2021ruite},
and conditional completion based on partially available elements~\cite{gupta2021layouttransformer},
\etc. However, none of them can cope with all these application scenarios simultaneously. This imposes a big challenge in unifying various layout generation subtasks with a single model, including conditional generation upon various specified attributes and unconditional generation from scratch.
\begin{figure}
    \centering
    \includegraphics[width=\columnwidth]{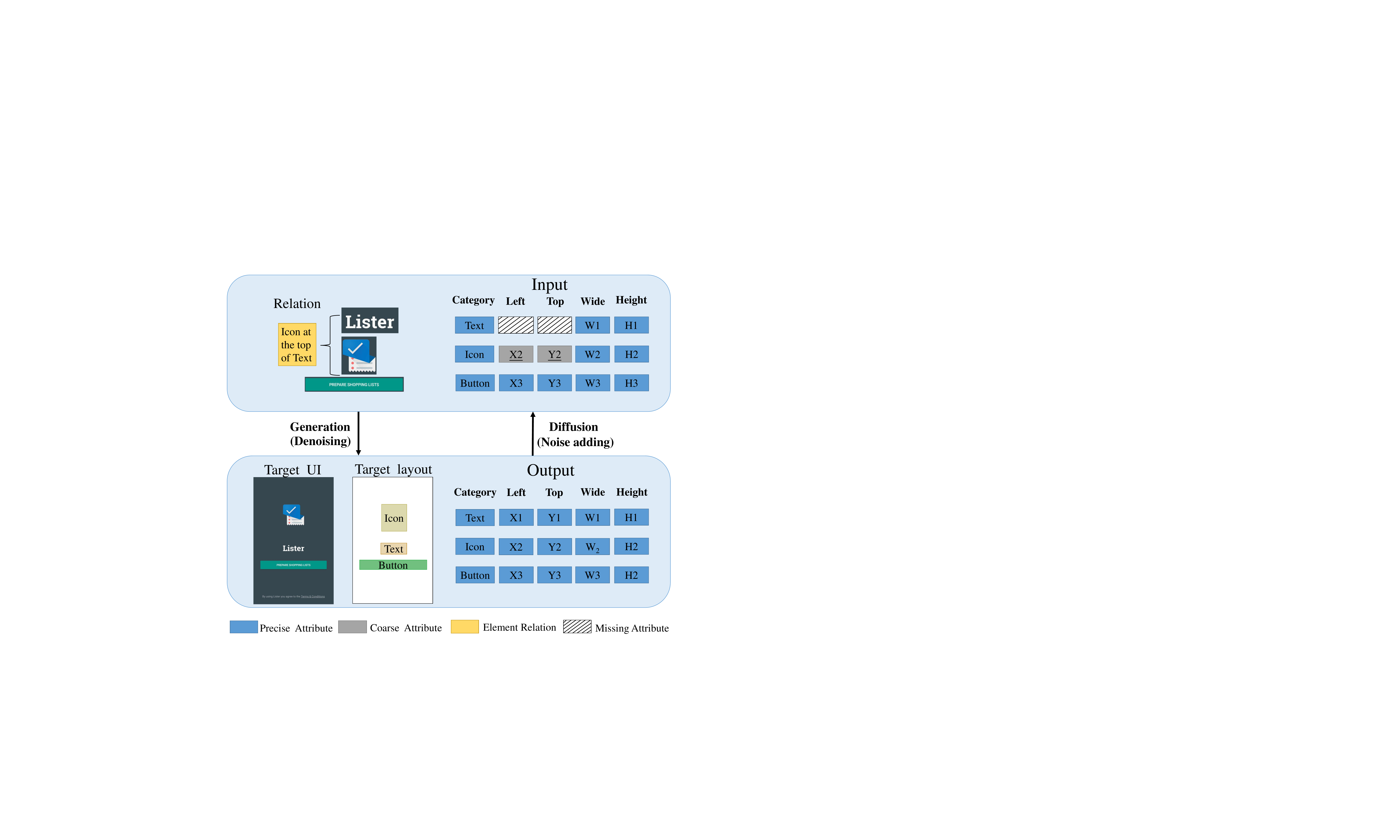}
    \caption{The layout generation tasks can be unified into a diffusion (noise-adding) process and a generation (denoising) process.}
    \label{fig:add/denoise}
    \vspace{-4mm}
\end{figure}
Towards this goal, the prior work UniLayout~\cite{jiang2022unilayout} takes a further step by proposing a multi-task framework to handle six subtasks for layout generation with a single model. However, the supported subtasks are pre-defined and could not cover all application scenarios, \egno, conditional generation based on specified element sizes. Besides, it does not take into account the combinational cases of several subtasks, \egno, the case wherein some elements have missing attributes to be generated while the others are with coarse attributes to be refined in the same layout simultaneously. 

\begin{figure}
    \centering
    \includegraphics[width=\columnwidth]{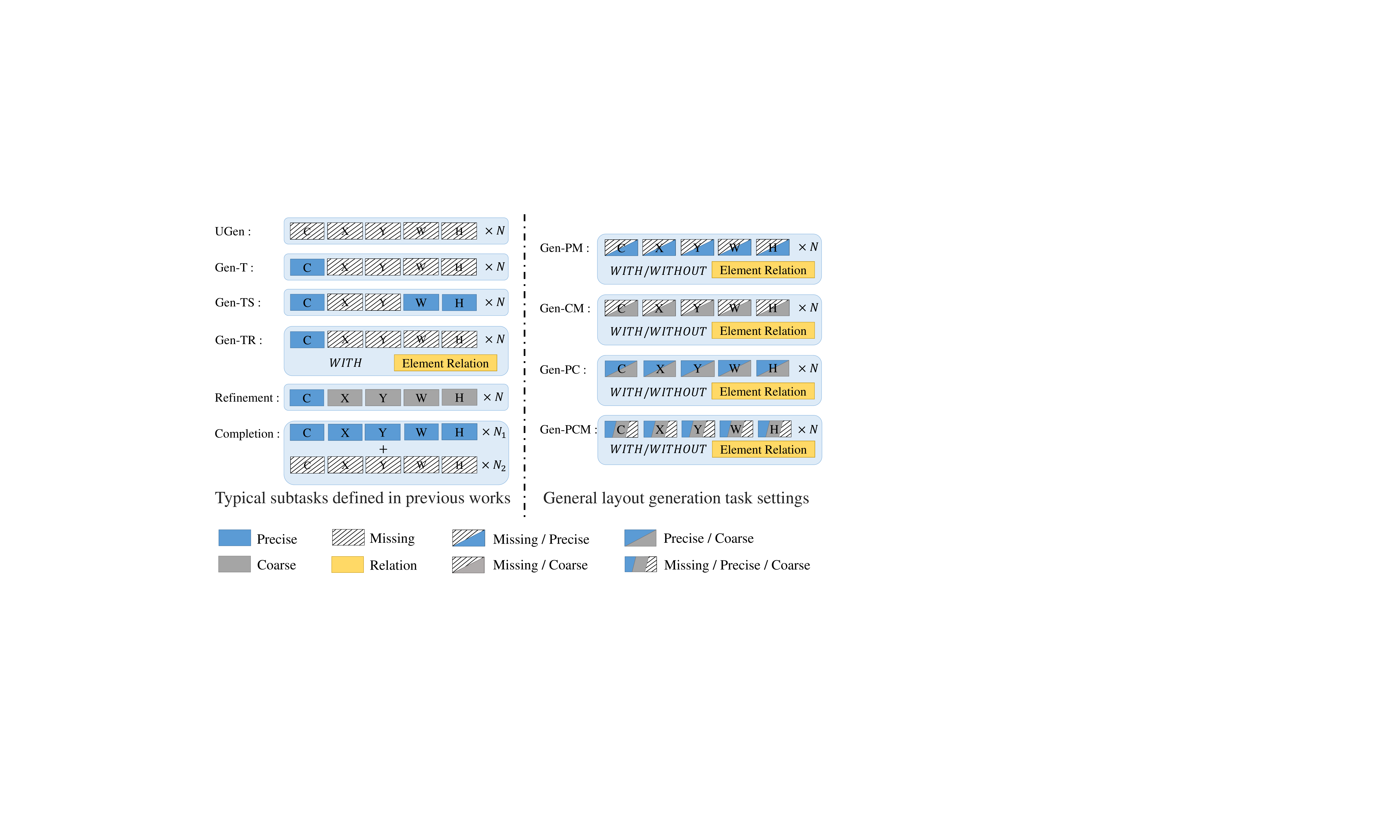}
    \caption{General task settings. The typical layout generation subtasks (left) can be covered by more general task definitions (right).}
    \label{fig:Typical Tasks}
    \vspace{-4mm}
\end{figure}

Generally, a layout comprises a series of elements with multiple attributes, \ieno, category, position, size and between-element relation. Each element attribute has three possible statuses: precise, coarse or missing. Different layout generation subtasks supported by previous works are defined as a limited number of cases where the attribute statuses are fixed upon attribute types, as shown in Figure~\ref{fig:Typical Tasks}. From a unified perspective, all missing or coarse attributes can be viewed as the corrupted results from their corresponding targets. With this key insight in mind, we innovatively propose to unify various forms of user inputs as intermediate statuses of a diffusion (corruption) process while modeling generation as a reverse (denoising) process.

Furthermore, attributes with different corruption degrees are likely to appear at once in user inputs. And different attributes have their own semantics and characteristics. These in fact impose a challenge for the diffusion process to create diverse training samples as comprehensive simulation for various user inputs. In this work, we propose a decoupled diffusion model \ours{} to address this challenge.
The meaning of ``decoupled'' here is twofold: \textit{(i)}~we design attribute-specific forward diffusion processes upon the attribute types; \textit{(ii)}~we decouple the forward diffusion process with the reverse denoising process, wherein the forward processes are individual for different types of attributes, whereas the reverse processes are integrated into one to be jointly performed.
In this way, our proposed \ours{} includes not only attribute-aware forward diffusion processes for different attributes to ensure the diversity of generation results, but also a joint denoising process with fully message passing over the global-scope elements for improving the generation quality. Our contributions can be summarized in the following:



\begin{itemize}[noitemsep,nolistsep,topsep=0pt,leftmargin=*]
    \item We present that
    various layout generation subtasks
    can be comprehensively unified with a single diffusion model.
    \item We propose the \oursfull{} (\ours{}), which allows parallel decoupled diffusion processes for different attributes and a joint denoising process for generation with sufficient global message passing and context exploitation. It conforms to the characteristics of layouts and achieves high generation qualities.
    \item Extensive qualitative and quantitative experiment results demonstrate that our proposed scheme outperforms existing layout generation models
    in terms of functionality and performance on different benchmark datasets.
\end{itemize}

\section{Related Works}
\label{sec:related}
\subsection{Layout Generation}
\label{subsec:related_lg}
Layout generation is a burgeoning research topic of synthesizing graphic scenes upon user requirements, facilitating manual design works in diverse applications. Early works in this area are commonly based on GAN~\cite{li2019layoutgan,li2020attribute,zheng2019content,kikuchi2021constrained} or VAE~\cite{patil2020read,jiang2022coarse,jyothi2019layoutvae,lee2020neural}. Recently, transformer models~\cite{gupta2021layouttransformer,arroyo2021variational} are emerging in this field to improve the generation diversity and quality. They are still difficult to achieve controllable generation since they predict attributes sequentially. To address this problem, BLT~\cite{kong2021blt} employs a bidirectional transformer to achieve parallel decoding.

In this field, the versatility across different generation subtasks is critical to make this technology practical in industry. Towards this goal, multi-task schemes~\cite{kong2021blt,jiang2022unilayout} are studied.
They are able to handle multiple subtasks simultaneously, but are limited to these pre-defined subtasks only. They do not consider the deep connection among various subtasks, and are thus unable to cover all task types in practical layout generation applications.
In this work, we study a versatile framework giving consideration to both the performance and functionality.

\subsection{Diffusion Models}
\label{subsec:related_diffusion}
Diffusion generative models~\cite{ho2020denoising,zhu2022discrete,nichol2021improved} have recently emerged as a new class of generative models of high performance. They use variational inference to produce training samples by adding noises until the signal is corrupted corresponding to a forward diffusion process, and learns to generate the signal through multi-step denoising corresponding to a reverse denoising process. It is firstly proposed
by Sohl-Dickstein \etal~\cite{sohl2015deep}
and regains widespread attention due to its rather impressive performance in generating images~\cite{ho2020denoising,song2021implicit,nichol2021improved,austin2021structured,gu2022vector}, texts~\cite{austin2021structured,li2022diffusion,gong2022diffuseq}, audio~\cite{wavegrad,kong2020diffwave,popov2021grad}, and more. In these works, the signal generation process is decomposed
into
multiple denoising steps where noises are added during training without distinction on different components/attributes of signals.
In this work, considering that elements in a layout include attributes that are of different semantics, we propose a decoupled diffusion model for layout generation to decouple these attributes in noise adding strategies.
It comprehensively unifies diverse generation subtasks with a single diffusion model.


\section{Problem Definition}
\label{sec:definition}


A layout $\bm{l}$ consisting of $N$ elements could be represented as a fully connected graph,
where the edges denote relations between elements. Each element has five attributes described by ($c$, $x$, $y$, $w$, $h$). $c$ stands for the category of each element such as the text, image, button, \etc. $(x, y)$ are the coordinates of the left-top corner of each element bounding box, denoting the information of location. $(w, h)$ describe the element size, corresponding to width and height, respectively. We denote pairwise relations between elements as
a matrix $\mathcal{E} \in \mathbb{R}^{N\times N}$.
As a result, such a layout could be formulated as $\bm{l} = [c_{1}, x_{1}, y_{1}, w_{1}, h_{1}, c_{2}, x_{2}, y_{2}, \cdots, h_{N}, \mathcal{E}]$.

Layout generation aims to predict all variables in $\bm{l}$ upon user requirements. For conditional layout generation, only partial attribute variables are available as conditions to generate the others. Unconditional layout generation requires the generation of all variables in $\bm{l}$ with only their total number $N$ given. Existing multi-task layout generators~\cite{kong2021blt,jiang2022unilayout} split the attribute variables in $\bm{l}$ into conditions and the ones to be predicted with rather limited number of protocols (3 in \cite{kong2021blt} while 6 in \cite{jiang2022unilayout}) according to the types of predefined subtasks. 
In this work, we make the first endeavour to eliminate this limitation towards comprehensive versatility.

\section{\oursfull}
\label{sec:method}

Our goal is to design a versatile framework that allows to take arbitrary attribute variables in $\bm{l}$ as conditions to predict the missing ones or refine the coarse ones. All elements after generation shall be able to be composed into a graphic layout that is functional and aesthetically pleasing. A big challenge for this lies in unifying multiple generation subtasks with a single model. Our key insight for addressing this is that \textit{the process from a completed layout to fully corruption can be modeled as a diffusion process, wherein partially available attribute variables in $\bm{l}$ can be viewed as corrupted results of the corresponding targets.}

With the above key insight, we propose \oursfull{} (\ours{}). Similar with prior diffusion generative models, \ours{} decomposes a generation process into successive denoising steps from noisy signals. It destroys training samples by successively adding noises to them, and then learns to recover them by reversing the noise addition process. Considering the characteristic of layout that different attributes have their own semantics in \ours{}, we innovatively propose decoupled diffusion processes with an attribute-specific noise-adding strategy and a joint reverse denoising process.

\subsection{Unification with Diffusion Modeling}
\label{sec:forward}

We pinpoint that a missing or coarse attribute in layouts could be viewed as the corrupted result of a complete one through a forward Markov diffusion process. In this section, we first give a unified formulation of the diffusion attributes and then elaborate our proposed decoupled corruption (noise-adding) strategy for different attributes.

For problem simplification, we quantize geometric attributes $x, y, w, h$ as integers following the common practices \cite{kong2021blt,jiang2022unilayout} in this field. So, attributes in $\bm{l}$ are all discrete variables. Like VQ-diffusion \cite{gu2022vector}, we model a discrete diffusion process with a status transition matrix. Given an attribute $x\in\left\{1,2,\cdots,K\right\}$ at time $t-1$, denoted by $x_{t-1}$, the probabilities that $x_{t-1}$ transits to $x_{t}$ could be represented by the matrix $\left[Q_t \right]_{ij}=q(x_t=i|x_{t-1}=j) \in \mathbb{R}^{K \times K}$. The forward Markov diffusion process can be formulated as:
\begin{equation} \label{eq:forward}
    q(x_t|x_{t-1}) = \bx^\top_{t}Q_t\bx_{t-1},
\end{equation}
where $\bx \in \mathbb{R}^{K \times 1}$ is the corresponded one-hot vector of $x$.
According to the property of Markov chains, the probability of $x_t$ from $x_0$ can be directly marginalized out as:
\begin{equation} \label{eq:prior}
    q(x_t | x_0) = \bx^\top_{t}\overline{Q}_{t}\bx_{0}, 
     \text{where}\;\overline{Q}_{t} = Q_1  Q_2 \cdots Q_t.
\end{equation}

Conditioned
on $x_0$, we can infer the posterior of this diffusion process by:
\begin{equation} \label{eq:posterior}
\begin{aligned}
    q(x_{t-1}|x_{t}, x_0) &= \frac{q(x_{t}|x_{t-1}, x_0)q(x_{t-1}|x_0)}{q(x_{t}|x_0)}\\
    &=\frac{
	\left(\bx^\top_{t} {Q}_t \bx_{t-1}\right) 
	\left(\bx^\top_{t-1} \overline{Q}_{t-1} \bx_0\right)}
	{\bx^\top_{t} \overline{Q}_t \bx_0}.
\end{aligned}    
\end{equation}


\vspace{-4mm}
\paragraph{Decoupled corruption (noise-adding) strategy.}

Layout is a graphic representation whose different attributes have their own semantics. A unified framework of the comprehensive versatility requires that the model can condition on any available precise attributes to generate or refine the remaining ones. To ensure the diversity of training samples, as shown in Algorithm \ref{alg:decoupled_diffusion}, we propose to decouple the entire forward diffusion process into three with their individual timelines for the attributes of category $c$, position $(x, y)$ and size $(w, h)$ and corrupt these three groups with different noises. Considering they are all discrete variables, similar to \cite{gu2022vector}, we adopt mask-and-replace strategies for their diffusion processes, which has a unified formulation as:
\begin{equation}
 \begin{split}
   Q_t =
   \begin{bmatrix}
   \alpha_t & \beta_t & \beta_t & \cdots & 0 \\
   \beta_t & \alpha_t & \beta_t & \cdots & 0 \\
   \beta_t & \beta_t & \alpha_t & \cdots & 0 \\
   \vdots & \vdots & \vdots & \ddots & \vdots \\
   \gamma_t & \gamma_t &\gamma_t  & \cdots & 1 \\
   \end{bmatrix},
 \end{split}
\label{eq:unified_strategy}
\end{equation}
where $\alpha_t, \beta_t, \gamma_t \in [0, 1]$.
In the diffusion process, the probabilities for each attribute variable to remain its original value, to be replaced with another value, and to be masked to be an absorbing status at the current time step are $\alpha_t$, $\beta_t$, and $\gamma_t$, respectively.
Such absorbing status is easy to be identified by networks and appears more as time $t$ increases, playing the role of embedding the temporal information about $t$ in each decoupled diffusion process. With this unified formulation, we adopt different instantiations for category attribute $c$ and geometry-related attributes $x, y, w, h$.



For category $c$, we adopt noises of a uniform distribution for its diffusion, corresponding to a transition matrix $Q_t^{c}$. In $Q_t^{c}$, $\alpha_t^c, \beta_t^c, \gamma_t^c$ are all constants for a given $t$ and satisfy that $\alpha_t^c + (K^c\!-\!1)\beta_t^c(1-\gamma_t^c) + \gamma_t^c = 1$. The $\beta_t^c$ and $\gamma_t^c$ increase linearly as time $t$ increases. Here, $K^c$ is the number of element categories in layouts, which varies for different datasets. (Detailed introduction is in the supplementary.)

For position $(x, y)$ and size $(w, h)$, we adopt discretized Gaussian noises \cite{austin2021structured} for their diffusion processes. Here, we introduce the formulation of their transition matrices, with the one for $h$ as an example. Other attribute variables follow the same formation with different value ranges. For $h$, the adopted discretized Gaussian noises correspond to a transition matrix $Q_t^{h}$. In $Q_t^{h}$, $\gamma_t^h$ is a scalar that increases linearly as time $t$ increases, and $\alpha_t^h$ and $\beta_t^h$ at the position $(i, j)$ are as below:
\begin{align}
[\alpha_t^h]_{ij} &= 1 - {\textstyle \sum_{j=0, j\neq i}^{K^h} [Q_t^h]_{ij}},\label{eq:gaussian_alpha} \\
[\beta_t^h]_{ij} &= \frac{(1-\gamma_t^h)\exp\left(-\frac{4|i - j|^2 }{(K^h-1)^2\sigma_t^h}\right)}{\sum_{n=-(K^h-1)}^{K^h-1}\exp\left(-\frac{4n^2 }{(K^h-1)^2\sigma_t^h}\right)}, \label{eq:gaussion_beta}
\end{align}
where $K^h$ is the number of values for $h$. 
And $\sigma_t^h$ is a linearly increasing hyper-parameter as time $t$ increases, which influences but is not equal to the variance of the discretized Gaussian noises.

\begin{algorithm}[!t]
    \caption{Training of the \ours}
    \label{alg:decoupled_diffusion}
    \begin{algorithmic}[1]
      \Require{Transition matrices $\{Q_t^c, Q_t^p, Q_t^s\}$, initial network parameters $\theta$, loss weight $\lambda$, and learning rate $\eta$.}
      \Repeat
      \State $l \gets $ sample a layout from the training set
      \State $timsteps= \text{zeros}(\text{len}(l)) $ \Comment{Record t of attributes.}
      \State {$\hat{l} = \text{RandSelect}(l)$} \Comment{Select attributes for diffusion.}
      \State {$\hat{l} = [C, P, S]$} \Comment{Group $\hat{l}$ upon the semantics.}
      \For {$g$ in $[C, P, S]$}                
          \State {sample $t \sim \text{Uniform}(\{1, \cdots, T\})$} 
          \For {$x$ in $g$}
          \State $timsteps[x.index] = t$
          \State $x = x_{t} \gets  \text{sample from}~q(x_t|x_0)$ \Comment{Eqn.~\ref{eq:prior} }
      \EndFor
      \EndFor
      \State $\mathcal{L}_x = \begin{cases}
            \lambda\mathcal{L}_{rec}, &\text{if } timsteps[x.index] =0 \\
			\mathcal{L}_0, &\text{if } timsteps[x.index] = 1\\
			\mathcal{L}_{t-1} , &\text{otherwise}
		\end{cases}$
      \State $\mathcal{L} = \sum_{x \in l}{\mathcal{L}_x}$
      \State $\theta \gets \theta - \eta \nabla_\theta \mathcal{L}$ \Comment{Update network parameters.}
      \Until{converged}
    \end{algorithmic}
\end{algorithm}

\subsection{Generation with a Joint Denoising Process}
\label{sec: reverse}

\begin{figure}
    \centering
    \includegraphics[width=\columnwidth]{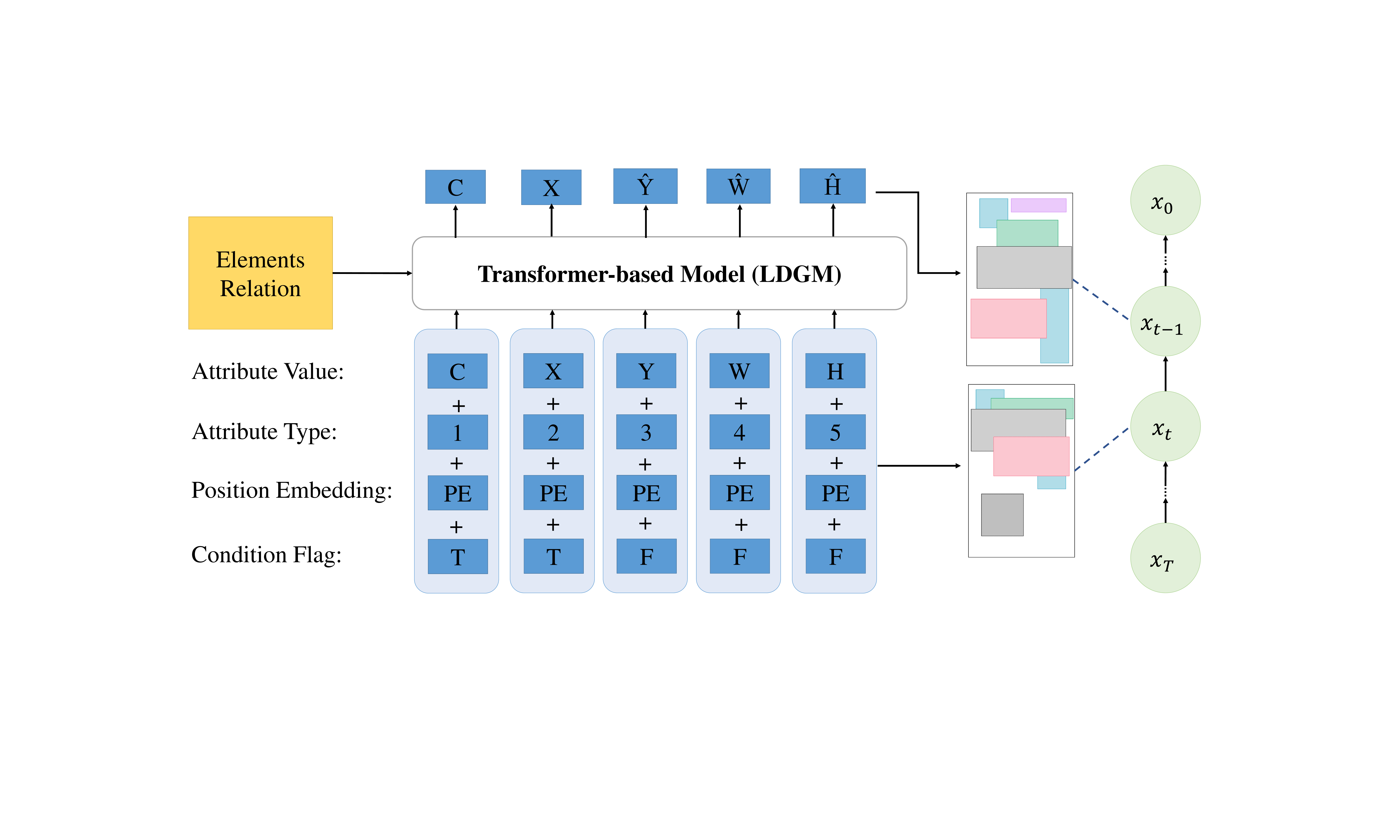}
    \caption{Overall framework of our method. The input attributes of different subtasks can be considered as different $x_t$. \ours{} gradually denoise them to $x_0$ as the final generation results.}
    \label{fig:Overall}
\end{figure}

Similar with other diffusion generative models \cite{ho2020denoising,austin2021structured,gu2022vector}, we train a denoising model as the generator to reverse the diffusion processes. In notable contrast to them, as presented in Algorithm \ref{alg:decoupled_diffusion}, our diffusion processes are decoupled for enhancing sample diversities, in which different types of attributes do not share a diffusion timeline. Layouts are highly structured representations. Thus, we propose a joint reverse denoising process for generation from scratch or corrupted layouts that consists of attributes with different degrees of corruption. Besides the exploitation of global-scope contexts, a joint reverse process enables the generation conditional on given relations between elements. To achieve these, we design a transformer-based model as shown in Figure~\ref{fig:Overall}. We introduce the formulation, model architecture and inference in the following.

Mathematically, the generator $p_\theta(x_{t-1}|x_t,\bm{g}(x_t))$ learns the reverse denoising process by estimating the transition posterior $q(\bx_{t-1}|\bx_t, \bx_0)$. The $\bm{g}(x_t)$ denotes the global-scope contexts of $x_t$ including other attributes in the current layout and the given relations between elements. This model is trained by optimizing the variational upper bound on corrupted (missing or coarse) attributes, \ieno, $\mathcal{L}_{vlb}\!=\!\mathbb E_{q(\bx_0)}\left[{\textstyle \sum_{t=0}^{T}\mathcal{L}_t}\right]$, and a reconstruction objective $\mathcal{L}_{rec}$ on precise attributes. The $\mathcal{L}_t$ in the variational upper bound can be detailed as:
\begin{equation}\label{eqn:L_vlb}
\small
    \mathcal{L}_t\!=\!\begin{cases}-\log p_\theta(x_0|x_1, \bm{g}(x_1)),&t\!=\!0\\ 
    D_{kl}({q(x_{t}|x_{t+1}, x_0)}||{p_\theta(x_t|x_{t+1},\! \bm{g}(x_{t+1}))}),&t\in[1,T)\\
    D_{kl}({q(x_T|x_0)}||{p(x_T)}),&t\!=\!T
    \end{cases}
\end{equation}
where $T$ is the maximum timestep in the diffusion process. Note that $p(\bm{x}_T)$ is the prior noise distribution that can be computed in advance during training. For the precise attributes without corruption, we adopt a reconstruction loss as below to ensure their preservation: 
\begin{equation}
\mathcal{L}_{rec} = -\log p_\theta(\hat{x}|x, \bm{g}(x)),
\label{eq:reconstruction}
\end{equation}
Where $\hat{x}$ refers to the output of our generative model for $x$. The overall loss is a weighted sum of $\mathcal{L}_{vlb}$ and $\lambda\mathcal{L}_{rec}$ with a hyperparameter $\lambda$:
\begin{equation}
\mathcal{L}= \mathcal{L}_{vlb}+\lambda\mathcal{L}_{rec}.
\end{equation}


\vspace{-4mm}
\paragraph{Model architecture.}

In \ours, as illustrated in Figure \ref{fig:Overall}, we adopt a transformer-based model to implement $p_\theta(x_{t-1}|x_t,\bm{g}(x_t))$. We tokenize each attribute of elements in the layout $\bm{l}$ with its relevant information including the \textit{value}, \textit{type}, \textit{position embedding} and \textit{condition flag}. Here, the type is an index to identify the category of this attribute while the position embedding is the embedding of element-level indexes indicating which layout element this attribute belongs to. The condition flag is a binary scalar to tell this attribute is precise or corrupted. All information is vectorized and then fused into one attribute token by a summation operation. Tokens in $\bm{l}$ are taken as the inputs of the generator $p_\theta$ to infer the denoising result for each attribute, in which a global-scope message passing over all layout elements and their attributes is performed via self-attention.

As introduced in
Section~\ref{sec:definition},
\ours{} supports the generation conditional on given relations. Given $N$ elements in layout $\bm{l}$, the pairwise relations can be represented by a matrix $\mathcal{E} \in \mathcal{R}^{N \times N}$. Each element in this matrix has nine possible discrete values, including three on size (\ieno, \textit{smaller}, \textit{larger}, and \textit{equal}), five on location (\ieno, \textit{above}, \textit{bottom}, \textit{left}, \textit{right}, and \textit{overlapped}) and another one to denote ``\textit{unavailable}''. We embed each value to be two vectors of the same dimension with the input token with two different embedding layers for query tokens and key tokens, respectively, yielding $V_{r}^K,V_{r}^Q\in \mathbb{R}^{N\times N\times d}$. We integrate such relation information into the generation process via relative position embedding proposed in \cite{shaw2018self}, formulated by:
\begin{equation} \label{eq: relation}
    e_{i,j}=\frac{(\boldsymbol{x}_iW^{Q}+[V_{r}^Q]_{i,j})(\boldsymbol{x}_jW^{K}+[V_{r}^K]_{i,j})}{\sqrt{d}},
\end{equation}
where $i$ and $j$ are token indexes, $\boldsymbol{x} \in \mathbb{R}^{1 \times d }$ is the vector of $x$. The final attention weight between these two tokens is:
\begin{equation}
    a_{i,j}=exp(e_{i,j})/{\textstyle \sum_{k=1}^{N}exp(e_{i,k})}.
\end{equation}


\vspace{-4mm}
\paragraph{Model inference.}

We propose a confidence-based inference strategy for \ours{} in
Algorithm~\ref{alg:Inference},
which can prevent generation errors from spreading across tokens via transformer in successive denoising steps. For missing attributes, their corresponding probabilities can be taken as confidence scores. At each denoising step, we merely preserve the predicted results of missing attributes with top-k high confidences and re-mark the remaining ones as absorbing status until all missing attributes are predicted. This operation is denoted by $\text{Top-}k\text{Keep}(\cdot)$ in
Algorithm~\ref{alg:Inference}
for brevity. For coarse attributes, they are continuously refined until the end of denoising. More details are in
Algorithm~\ref{alg:Inference}.

\begin{algorithm}[!t]
    \caption{Inference of the \ours{}}
    \label{alg:Inference}
    \begin{algorithmic}[1]
      \Require{Initial layout $\bm{l}_T$, condition flags, and maximum denoising steps $T$.}
      \State {$\bm{l}_T \gets \text{tokenize } \bm{l}_T \text{with condition flags}$}
      \State {$\bm{l}_T^m \gets \text{GetMiss}(\bm{l}_T)$} \Comment{Get missing attributes from $\bm{l}_T$.}
      \State {$N_{m} \gets len(\bm{l}_T^m)$}
      \State {$k \gets \lceil N_{m}/T \rceil$}
      \For{$t = T, \cdots, 1$}
            \State {$p_\theta(\bm{l}_{t-1}|\bm{l}_t) = \ours{}(\bm{l}_t)$} 
            \State {$\bm{l}_{t-1}, \bm{p}_{t-1} \gets$ sample from $p_\theta(\bm{l}_{t-1}|\bm{l}_t)$ } 
        \If{$N_{m} > 0$} 
            \State {$\bm{l}_{t-1}^m, \bm{p}_{t-1}^m \gets \text{GetMiss}(\bm{l}_{t-1}, \bm{p}_{t-1})$}
            \State {$\bm{l}_{t-1}^m \gets \text{Top-}k\text{Keep}(\bm{l}_{t-1}^m, \bm{p}_{t-1}^m)$} 
            \State {$N_{m} \gets N_{m} - k$}
        \EndIf
      \EndFor
      \State \textbf{return} $\bm{l}_0$
    \end{algorithmic}
\end{algorithm}
          


\section{Experiments}
\label{sec:exp}

\subsection{Experiment Setup}
\label{sec:experiment_setup}

\paragraph{Datasets.}
We conduct ablation and comparison experiments on three public datasets, \ieno, Magazine~\cite{zheng2019content}, Rico~\cite{deka2017rico} and PubLayNet~\cite{zhong2019publaynet}. Magazine~\cite{zheng2019content} is a dataset of magazine pages with 6 layout element categories, containing 4K+ images. Rico~\cite{deka2017rico} contains 66K+ images of UIs for mobile applications with 27 element categories. PubLayNet~\cite{zhong2019publaynet} comprises 360K+ machine annotated document images with 5 element categories.
Following the common practices in previous studies~\cite{kikuchi2021constrained,lee2020neural,li2019layoutgan}, we clean the datasets to improve the quality of the datasets. For Rico dataset, only 13 most frequent categories are remained and the elements out of these categories are removed from the dataset. For both Rico and PubLayNet datasets, we remove the samples containing more than 25 elements.
Since the splitting protocols for training and testing are not consistent over different publications, we re-implement their proposed methods and report evaluation results with the same data splitting protocol for fair comparison in the following sections. Detailed introduction for datasets and their configurations can be found in our supplementary.

\vspace{-4mm}
\paragraph{Evaluation metrics.}
We adopt four widely-used evaluation metrics ($\uparrow$: the bigger the better. $\downarrow$: the smaller the better). They are introduced in the following.
\textit{Maximum Intersection-over-Union (MaxIoU)} $(\uparrow)$~\cite{kikuchi2021constrained} measures the similarity of the elements in bounding boxes of the same category label between the collections of generated layouts and ground-truth layouts.
\textit{Frechet Inception Distance (FID)} $(\downarrow)$ measures the distributional distance between the feature representations of generated layouts and their ground truth. 
Following~\cite{kikuchi2021constrained}, we train a model to classify whether the input layout is corrupted or not,
and use the output of the penultimate layer for FID computation.
\textit{Alignment $(\downarrow)$}~\cite{li2020attribute} is used to measure the alignment of elements in generated layouts for aesthetics assessment.
We compute this metric with respect to six items: \textit{left border, center at x-axis, right border, top border, center at y-axis, and bottom border}.
\textit{Overlap $(\downarrow)$}~\cite{li2020attribute} measures the overlapping degrees between each element pair inside generated layouts. A well-designed layout typically has less element overlaps.


\vspace{-4mm}
\paragraph{Evaluation subtasks.}

We evaluate our \ours{} on six existing layout generation subtasks previously defined in \cite{rahman2021ruite,gupta2021layouttransformer,kong2021blt,jiang2022unilayout} for performance comparison. Besides, we unify them into four more general settings and evaluate \ours{} on these settings to demonstrate our proposed scheme can provide more comprehensive versatility.
\begin{itemize}[noitemsep,nolistsep,topsep=0pt,leftmargin=*]
    \item \textit{Unconditional generation (U-Gen)} aims at generating 
    a layout with no input conditions provided by users.
    \item \textit{Generation conditioned on types (Gen-T)} is to generate 
    a layout conditioned on specified element types.
    \item \textit{Generation conditioned on types and sizes (Gen-TS)} generates 
    a layout with specified element types and sizes.
    \item \textit{Generation conditioned on types and relations (Gen-TR)} generates 
    a layout conditioned on specified element types and pairwise element relations.\footnote{Like 
    CLG-LO~\cite{kikuchi2021constrained}, we randomly sample 10\% relations as the inputs.}
    \item \textit{Refinement} 
    updates coarse attributes of elements in a layout to be more reasonable and realistic.
    \footnote{Following 
    RUITE~\cite{rahman2021ruite}, we synthesize an input layout by turning the precise attributes in real layout into coarse attributes, with the noise sampled from a normal distribution (mean: 0, standard deviation: 0.01).}
    \item \textit{Completion} aims at generating the missing attributes in a logout from the given/specified ones.
\end{itemize}

As we discuss in Section \ref{sec:definition}, each element in a layout can be described with five attributes, \ieno, $(c, x, y, h, w)$. Each attribute has three possible statuses in total: precise (P), coarse (C), or missing (M). When two of these three statuses may appear for any attribute, there are three combined settings, \ieno, \textit{Gen-PM}, \textit{Gen-CM}, and \textit{Gen-PC}. When these three statuses are all allowed, it corresponds to a most general setting, \ieno, \textit{Gen-PCM}. Note that all subtasks including six previously defined ones, \textit{Gen-PM}, \textit{Gen-CM}, and \textit{Gen-PC} can be viewed as the instantiations of \textit{Gen-PCM}. 
\vspace{-4mm}
\paragraph{Implementation detail.}

We set both maximum diffusion steps and denoising steps to $100$. For the network in \ours{}, we use 8 eight-head transformer layers. The embedding dimension is and the feed-forward dimension is 2048.
We implement our proposed \ours{} with PyTorch~\cite{paszke2019pytorch} and adopt the
AdamW optimizer~\cite{loshchilov2017decoupled} with $\beta_{1}=0.9$ and $\beta_{2}=0.98$ for model training on NVIDIA V100 GPUs. The batch size is set to 128, and the learning rate is 5e-5. Linear warmup schedule is adopted and the warmup proportion is set to 0.1. The $\lambda$ for loss weighting is set to 0.1. More implementation details (\egno, hyperparameter configurations) are in our supplementary.
\begin{table*}
    \centering
    \caption{Experiment results on different layout generation subtasks. Align. denotes the alignment metric.}
    \renewcommand{\arraystretch}{1.2}
    \begin{small}
        \resizebox{\textwidth}{!}{
            \begin{tabular}{llcccccccccccc}
                \toprule
                   \multirow{2}{*}{Subtasks}                             &    \multirow{2}{*}{Methods}                   & \multicolumn{4}{c}{Magazine}                                                                                         & \multicolumn{4}{c}{Rico}             & \multicolumn{4}{c}{PubLayNet}                                                                                                                                                                                                      \\ \cmidrule(l){3-6}           \cmidrule(l){7-10}           \cmidrule(l){11-14}
                                & & MaxIoU  $\uparrow$        & FID  $\downarrow$             & Align.  $\downarrow$       & Overlap    $\downarrow$    & MaxIoU $\uparrow$         & FID  $\downarrow$           & Align.  $\downarrow$       & Overlap  $\downarrow$       & MaxIoU $\uparrow$         & FID  $\downarrow$           & Align. $\downarrow$       & Overlap  $\downarrow$\\ \hline
                \multirow{4}{*}{U-Gen}       & LayoutTrans.~\cite{gupta2021layouttransformer}     & 0.18  & 47.84 & 0.59 & 47.98 & 0.46 & 46.64  & 0.66  & 64.10  & 0.32   & 49.72 & 0.37  & 36.63 \\ 
                                            & BLT~\cite{kong2021blt}                   & 0.20  & 44.91 & 0.55 & 55.56 & 0.51 & 33.81  & 0.59  & 67.33 & 0.34   & 48.24 & 0.27  & 42.79\\ 
                                            & UniLayout~\cite{jiang2022unilayout}             & 0.31 & 36.61 & 0.49& \textbf{44.50}   &\textbf{0.62} & 26.68 & 0.40 & 59.26     & 0.33 & 32.29 & \textbf{0.22} & 22.19\\ 
                                            & \ours{} (Ours)            & \textbf{0.38} & \textbf{32.73} & \textbf{0.47} & 46.43  & \textbf{0.62} & \textbf{26.06} & \textbf{0.36} & \textbf{56.35}     & \textbf{0.46} & \textbf{25.94} & 0.25 & \textbf{19.83}\\ \hline

                \multirow{4}{*}{Gen-T}      & LayoutGAN++ ~\cite{kikuchi2021constrained}          & 0.26 & 36.35 & 0.54 & 58.44   & 0.46 & 34.43 & 0.58 & 59.85     & 0.36 & 30.48 & 0.19 & 32.80 \\ 
                                            & BLT~\cite{kong2021blt}                   & 0.22 & 48.26 & 0.69 & 64.01   & 0.44 & 39.64 & 0.57 & 56.83     & 0.37 & 44.86 & 0.21 & 38.21\\ 
                                            & UniLayout~\cite{jiang2022unilayout}             & 0.32 & 28.37 & 0.51 & 53.56   & 0.55 & 18.06 & 0.48 & 57.92     & 0.41 & 27.34 & 0.20 & 20.98 \\ 
                                            & \ours{} (Ours)            & \textbf{0.36} & \textbf{24.67} & \textbf{0.45} & \textbf{45.11}   & \textbf{0.58} & \textbf{16.64} & \textbf{0.39} & \textbf{55.87}    & \textbf{0.44} & \textbf{20.69} & \textbf{0.15} & \textbf{16.88} \\ \hline

                \multirow{3}{*}{Gen-TS}     & BLT~\cite{kong2021blt}                   & 0.33 & 22.72 & 0.59 & 61.94   & 0.51 & 42.88 & 0.46 & 57.74    & 0.40 & 24.32 & 0.16 & 31.06    \\ 
                                            & UniLayout~\cite{jiang2022unilayout}             & 0.35 & 19.35 & 0.58 & 56.43   & 0.55 & 20.42 & 0.49 & 58.72     & 0.43 & 27.47 & \textbf{0.16} & 23.82  \\ 
                                            & \ours{} (Ours)            & \textbf{0.37} & \textbf{17.65} & \textbf{0.45} & \textbf{44.25}   & \textbf{0.62} & \textbf{12.59} & \textbf{0.35} & \textbf{55.92}    & \textbf{0.47}& \textbf{19.02} & \textbf{0.16} & \textbf{10.09}  \\ \hline

                \multirow{3}{*}{Gen-TR}      & CLG-LO~\cite{kikuchi2021constrained}                 & 0.27 & 33.88 & 0.59 & 59.43   & 0.38 & 38.89 & 0.54 & \textbf{56.51}     & 0.38 & 31.87 & 0.21 & 34.39   \\ 
                                            & UniLayout~\cite{jiang2022unilayout}             & 0.36 & \textbf{19.24} & 0.54 & 49.61   & 0.57 & 26.38 & 0.46 & 66.93     & \textbf{0.46} & 27.73 & 0.17 & 27.35    \\ 
                                            & \ours{} (Ours)            & \textbf{0.39} & 20.58 & \textbf{0.48} & \textbf{47.27}   & \textbf{0.61} & \textbf{16.98} & \textbf{0.39} & 58.75    & 0.44 & \textbf{19.54} & \textbf{0.16} & \textbf{21.28}  \\ \hline

                \multirow{3}{*}{Refinement} & RUITE~\cite{rahman2021ruite}                 & 0.24 & 44.27& 0.64 & 54.26 & 0.46 & 36.70 & 0.57 & 64.13     & 0.32 & 41.72 & 0.49 & 35.74   \\ 
                                            & UniLayout~\cite{jiang2022unilayout}               & 0.33 & 19.78 & 0.49 & 49.02   & 0.56 & 24.41 & 0.42 & 56.04    & 0.44 & 22.34 & 0.11 & 27.23 \\ 
                                                & \ours{} (Ours)            & \textbf{0.39} & \textbf{14.95} & \textbf{0.42} & \textbf{37.22}   & \textbf{0.62} & \textbf{13.19} & \textbf{0.33} & \textbf{52.17}    & \textbf{0.48} & \textbf{15.28} & \textbf{0.10} & \textbf{13.05}  \\ \hline

                \multirow{3}{*}{Completion} & LayoutTrans.~\cite{gupta2021layouttransformer}     & 0.17 & 39.36 & 0.67 & 55.32   & 0.46 & 36.15 & 0.66 & 67.10     & 0.32 & 41.72 & 0.37 & 39.81 \\ 
                                            & UniLayout~\cite{jiang2022unilayout}             & 0.23 & 28.78 & 0.52 & 46.43   & 0.59 & 25.18 & 0.45 & 55.99     & 0.41 & 32.04 &0.19 & 22.90  \\ 
                                            & \ours{} (Ours)           & \textbf{0.38} & \textbf{24.35} & \textbf{0.49} & \textbf{39.26}  & \textbf{0.60} & \textbf{16.42} & \textbf{0.36} & \textbf{53.15}     & \textbf{0.44} & \textbf{25.31} & \textbf{0.10} & \textbf{19.45}  \\ \hline
                Gen-PM      &  \multirow{4}{*}{\ours{} (Ours)}         & 0.38 & 27.33 & 0.47& 39.02 & 0.58 & 21.64 & 0.38 & 56.56    & 0.46 & 23.58 & 0.10 & 14.11 \\
                Gen-CM      &           & 0.37 & 28.74 & 0.51& 43.25 & 0.57 & 26.15 & 0.38 & 57.74    & 0.44 & 24.94 & 0.11 & 16.26 \\
                Gen-PC      &            & 0.37 & 22.56 & 0.47& 42.95 & 0.60 & 18.13 & 0.36 & 53.67    & 0.50 & 16.42 & 0.09 & 12.51 \\
                Gen-PCM      &            & 0.37 & 24.45 & 0.49& 44.41 & 0.59 & 21.59 & 0.40 & 54.77    & 0.42 & 25.76 & 0.14 & 19.68 \\ \hline
                GT         &-  & 0.41 & 9.89 & 0.43 & 34.27 & 0.66 & 7.05 & 0.26 & 49.86  & 0.64 & 9.38 & 0.008 & 5.18      \\                     
                \specialrule{1.1pt}{1pt}{0pt}
            \end{tabular}}                     
    \end{small}
    \label{Tab:quantitative_results}
\end{table*}

\subsection{Quantitative Results}




We compare our proposed \ours{} to state-of-the-art (SOTA) methods on the six previously defined subtasks introduced in Section \ref{sec:experiment_setup} to demonstrate its performance superiority. In addition, we also evaluate \ours{} on our newly proposed task settings to show the extended functionality towards comprehensive versatility. 
For fair comparison and convincing evaluation, we generate 1K layouts for each model and report the result averaged over five runs with random seeds for each experiment. The results are shown in Table~\ref{Tab:quantitative_results}. Their standard deviations are placed in the supplementary. 


As can be seen from Table~\ref{Tab:quantitative_results}, \ours{} achieves superior performance to SOTA layout generators across the three public datasets. It suggests that by unifying various layout generation subtasks with diffusion modeling, our proposed \ours{} is impressively effective in simultaneously handling multiple layout generation subtasks. Moreover, the quantitative results on newly proposed task settings demonstrate that our \ours{} is able to achieve more comprehensive versatility. We thus believe that it can support more diverse user requirements in practical applications.



\begin{figure*}[!t]
    \centering
    \includegraphics[width=\linewidth]{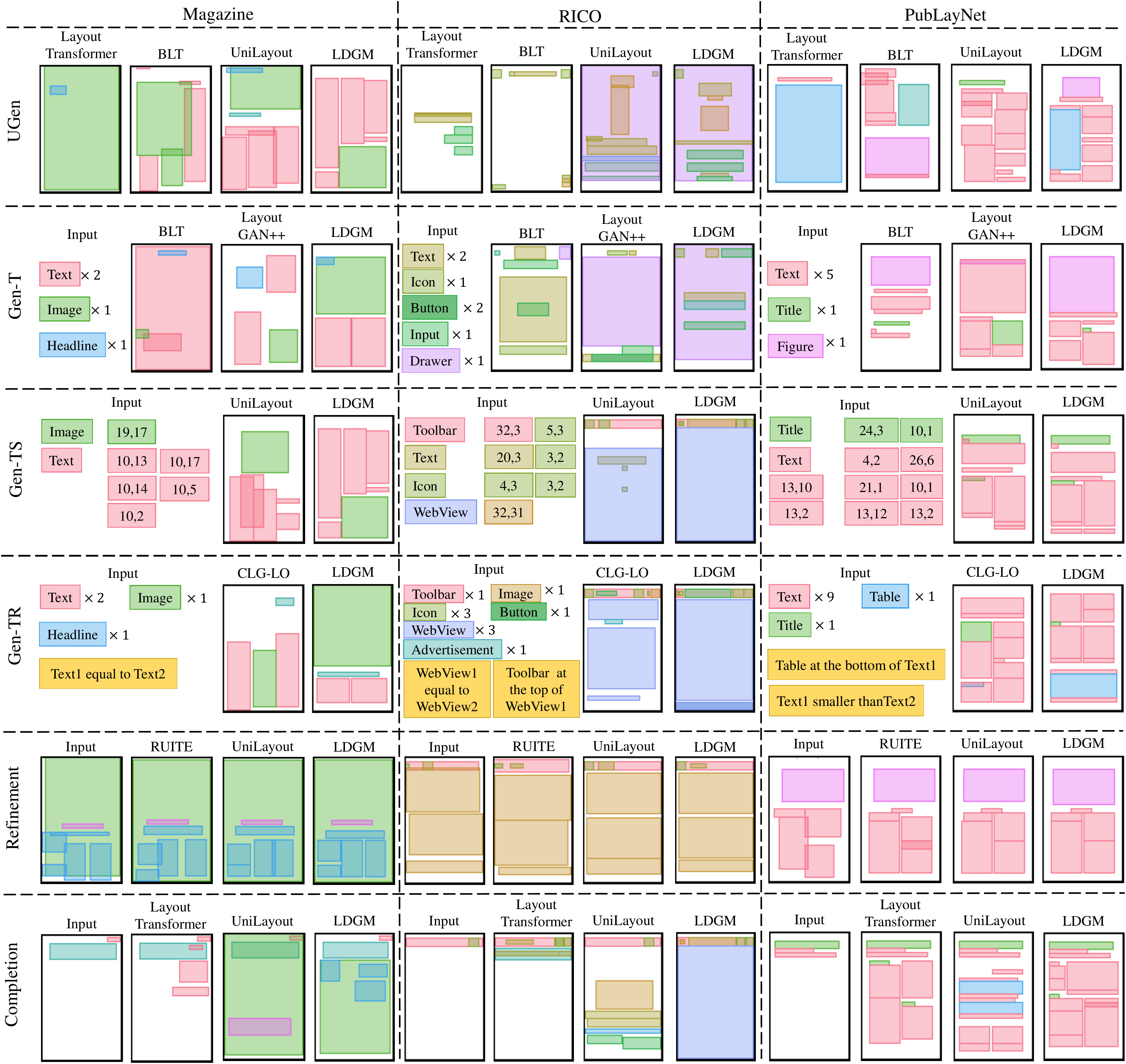}
    \caption{Qualitative comparisons with state-of-the-art layout generation methods.} 
    \label{fig:Qualitative}
\end{figure*}

\subsection{Qualitative Comparisons}


We compare visualizations of generated layouts from our \ours{} and other recent layout generators in Figure~\ref{fig:Qualitative}. It can be observed that \ours{} achieves superior generation performance to others with better alignment, less overlaps, and more realistic details. More visualization results, rendered images and analysis can be found in our supplementary.


\subsection{Ablation Studies}
\label{sec:ablation}


To validate the effectiveness of our proposed technical components in \ours{}, we conduct a series of ablation studies on the most general generation task \textit{Gen-PCM}. 

\vspace{-4mm}
\paragraph{Decoupled corruption strategy.}





Corruption strategy is crucial for diffusion models \cite{austin2021structured}, which is indiscriminate for different components of signals in previous works while \ours{} adopts an attribute-decoupled corruption strategy. We demonstrate the effectiveness of this design by comparing four different diffusion strategies\footnote{More details and illustrations are in our supplementary.}:
\textit{(i)}~\textit{Non-decoupled strategy}: adding noises for different types of attributes with a shared diffusion timeline. 
\textit{(ii)}~\textit{Partial-decoupled strategy}: the three types of attributes (\ieno, category, size and position) are involved in diffusion processes in order with three individual and partially overlapped diffusion timelines.
\textit{(iii)}~\textit{Sequential-decoupled strategy}: adding noises for three types of attributes sequentially with three individual and non-overlapped diffusion timelines.
\textit{(iv)}~our~\textit{Parallel-decoupled strategy (ours)}: adding noises for three types of attributes in parallel with three individual and fully overlapped diffusion timelines. The comparison results are in Table \ref{tab: Corruption Strategy}. We can observe that our proposed strategy achieves the best \textit{MaxIOU}, \textit{FID} and \textit{Alignment} compared to the other three. It delivers the second best \textit{Overlap} (very close to the best one) since the training samples of this strategy are of the highest diversity thus imposing the largest difficult for model optimization.

\begin{table}[!t]
    \caption{Experiment results on the Rico dataset by varying the corruption strategies on input tokens.}
    \label{tab: Corruption Strategy}
    \small
    \centering
    \resizebox{\linewidth}{!}{
    \begin{tabular}{lcccc}
      \toprule
      \multicolumn{1}{c}{Model}  &  MaxIoU  $\uparrow$        & FID  $\downarrow$             & Align.  $\downarrow$       & Overlap    $\downarrow$ \\
      \midrule
      Non-decoupled  & 0.56 & 29.24 & 0.43 & 60.04  \\ 
      Partial  &  0.57 & 27.71 & 0.48 & \textbf{54.24}  \\
      Sequential  &  0.56 & 26.69 & 0.43 & 57.17  \\
      Parallel (Ours)  & \textbf{0.59}	&\textbf{21.59}&	\textbf{0.40}	&54.77 \\ 
      \bottomrule
    \end{tabular}
    }
\end{table}

\vspace{-4mm}
\paragraph{Inference strategy.}


We compare our inference strategy against two common transformer decoding strategies: \textit{(i)}~\textit{Autoregressive}: the tokens are decoded one by one in the sequence order. \textit{(ii)}~\textit{Non-autoregressive}: where tokens are decoded
all at once for each time step for $T$ decoding steps in total.
As shown in Table~\ref{tab: Inference Strategy},
\textit{Non-autoregressive} achieves the most inferior performance, which is because it predicts many missing attributes at one time step, leading to propagation of generation errors across tokens. 
\textit{Autoregressive} requires to decode tokens in a given order, which means it lacks the flexibility to meet user requirements as discussed in \cite{kong2021blt}.
Our proposed strategy \ours{} can adaptively decode tokens starting from easier ones based on the confidence scores, enabling the exploitation of more reliable context information when decoding the harder ones.

\begin{table}[!t]
    \caption{Experiment results on the Rico dataset in terms of different inference strategies.}
    \label{tab: Inference Strategy}
    \small
    \centering
    \resizebox{\linewidth}{!}{
    \begin{tabular}{lccccc}
      \toprule
     \multicolumn{1}{c}{Model}  &  MaxIoU  $\uparrow$        & FID  $\downarrow$             & Align.  $\downarrow$       & Overlap    $\downarrow$  \\
      \midrule
      AutoReg   &\textbf{0.60}&  23.16&  0.42& 56.87   \\ 
      Non-AutoReg  &0.57& 25.14 & 0.44 &  58.63 \\ 

      Ours  & 0.59	&\textbf{21.59}&	\textbf{0.40}	&\textbf{54.77} \\ 

      \bottomrule
    \end{tabular}
    }
\end{table}

\begin{table}[!t]
    \caption{Ablation study on condition flags. \textit{Retent.} refers to the retention/unchanged ratio of condition attributes free of generative errors. ``w/o. C-Flags'' denotes discarding condition flags.}
    \label{tab: Condition Flags}
    \small
    \centering
    \resizebox{\linewidth}{!}{
    \begin{tabular}{lccccc}
    \toprule
    \multicolumn{1}{c}{Model}  &  MaxIoU  $\uparrow$        & FID  $\downarrow$             & Align.  $\downarrow$       & Overlap    $\downarrow$ & Retent.    $\uparrow$\\
    \midrule
    w/o. C-Flags   &0.55&  27.38&  0.42& 55.54 & 11.25  \\ 
    Ours  & \textbf{0.59}	&\textbf{21.59} & \textbf{0.40}	& \textbf{54.77} & \textbf{99.66}\\
    \bottomrule
    \end{tabular}
    }
\end{table}

\vspace{-4mm}
\paragraph{Condition flags.}




We study the benefits of condition flags by comparing \ours{} to the model without them. As presented in Table~\ref{tab: Condition Flags}, the use of condition flags not only improves the generation qualities measured by different metrics but also plays an important role in preventing attributes given as conditions from being destroyed during generation.

\section{Conclusion and Future Work}
In this work, we unify diverse layout generation subtasks, including unconditional generation from scratch and conditional generation based on various user inputs, with a single diffusion model, \ieno, \oursfull{} (\ours). Given the number of elements in the layout, \ours{} supports the generation from arbitrary available element attributes, including category, position, size and relation between elements, no matter they are coarse or precise. To the best of our knowledge, this is the first endeavour to achieve such a comprehensively versatile layout generator. Besides, we devise a decoupled diffusion model which performs decoupled diffusion processes for attributes of different semantics/characteristics and generates them jointly with global-scope contexts taken into account. This methodology presents a core idea of ``decouple-first-diffusion-then''. We believe this idea will inspire more exploration in designing diffusion-based generative models.

{\small
\bibliographystyle{ieee_fullname}
\bibliography{main}
}

\clearpage
\section*{\Large{\textbf{Supplementary Material}}}
In this supplementary material, we detail the experimental setup in Section~\ref{sec:supp_exp_setup}, including detailed introduction to datasets in Section~\ref{subsec:datasets}, hyperparameter configurations in Section~\ref{subsec:hyperparameter}, and more implementation details of ablation experiments in Section~\ref{subsec:implementation_details}. Moreover, we provide more experimental results in Section~\ref{sec:more_exp}, including \textit{(i)}~ablation studies for noise types and decoupling levels in Sections~\ref{subsec:ablation_noise}, \ref{subsec:ablation_decoupling}, respectively, and \textit{(ii)}~more quantitative and qualitative results in Sections~\ref{subsec:more_quantitative} and \ref{subsec:more_qualitative}, respectively.

\section{More Details about Experiment Setup}
\label{sec:supp_exp_setup}

\subsection{Detailed Introduction to the Datasets}
\label{subsec:datasets}

Three datasets, \ieno, Magazine~\cite{zheng2019content}, Rico~\cite{deka2017rico}, and PubLayNet~\cite{zhong2019publaynet}, are adopted in our experiments. Since the data splitting protocols for training and testing are not consistent over different publications, we thus re-implement the proposed methods in them and report the experiment results with the same data
splitting protocols introduced below for fair comparison.  
\begin{itemize}[noitemsep,nolistsep,topsep=0pt,leftmargin=*]
    \item \textit{Magazine}~\cite{zheng2019content} contains total $4$K+ images of magazine pages. We use 85\% of the dataset for training, 5\% for validation, and 10\% for testing.
    The categories in the dataset include \textit{text}, \textit{image}, \textit{headline}, \textit{text-over-image}, \textit{headline-over-image}, and \textit{background}.
    \item \textit{Rico}~\cite{deka2017rico} is a dataset of mobile app UI that contains $66$K+ UI layouts. We randomly select 85\% of the dataset for training, 5\% for validation, and 10\% for testing. Following the common practices in previous works~\cite{kong2021blt,jiang2022unilayout,kikuchi2021constrained}, we exclude elements whose labels are not in the 13 most frequent labels from using.
    The adopted categorise are \textit{Toolbar}, \textit{Image}, \textit{Text}, \textit{Icon}, \textit{Text Button}, \textit{Input}, \textit{List Item}, \textit{Advertisement}, \textit{Pager Indicator}, \textit{Web View}, \textit{Background Image}, and \textit{Drawer,Modal}. Following the common practices in previous works~\cite{lee2020neural,li2019layoutgan,kikuchi2021constrained}, we also exclude the layouts with more than 25 elements since these layouts are rare but may lead to low training efficiency.
    \item \textit{PubLayNet}~\cite{zhong2019publaynet} contains $360$K+ document layout examples crawled from the Internet. We adopt the full official training split for training, 33\% of the official validation split for validation, and the rest of the validation split for testing.
    This dataset contains elements from 5 catgories, \egno, \textit{Text}, \textit{Title}, \textit{List}, \textit{Table}, and \textit{Figure}. Following the common practices in previous works~\cite{lee2020neural,li2019layoutgan,kikuchi2021constrained}, similar to that for Rico, we exclude the layouts with more than 25 elements for improving the training efficiency. 
 \end{itemize}

\subsection{Hyperparameter Configurations}
\label{subsec:hyperparameter}

In our proposed \ours{}, the hyperparameter $\beta_t^c$ controls the transition probability from a category to the other one, which increases from $0$ to $0.02/K^c$ as time $t$ increases from $0$ to $T$. Here, $K^c$ is the number of adopted element categories in layouts, which varies for different datasets. For attributes $x,y,h,w$, $\sigma_t$ controls the transition probability from one discrete to the other. $\sigma_t^x$, $\sigma_t^y$, $\sigma_t^h$ and $\sigma_t^w$ increase from $0$ to $0.02$ as time $t$ increases from $0$ to $T$. For all attributes $c,x,y,h,w$, the hyperparameter $\gamma_t$ denotes the probability of masking a discrete value as an absorbing status. $\gamma_t^c$, $\gamma_t^x$, $\gamma_t^y$, $\gamma_t^h$ and $\gamma_t^w$ increase from $0$ to $0.032$ as time $t$ increases from $0$ to $T$. We adopt the same hyperparameter configurations in all experiments for fair comparison.


\subsection{Implementation Details of Ablation Study on Decoupled Corruption Strategy}
\label{subsec:implementation_details}
\begin{figure*}[htbp]
\centering
\begin{subfigure}[b]{0.3\textwidth}
   \includegraphics[height=0.13\textheight,width=\textwidth]{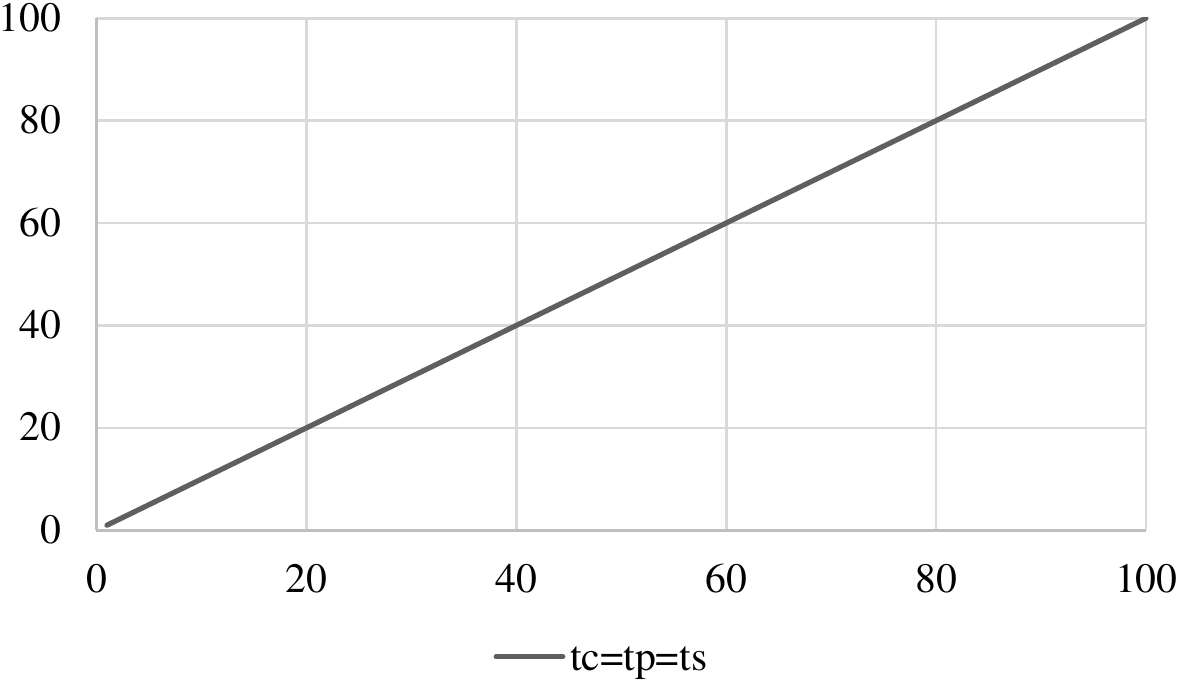}
   \caption{Non-decoupled strategy}
   \label{fig:Non-decoupled strategy} 
\end{subfigure}
\begin{subfigure}[b]{0.3\textwidth}
   \includegraphics[height=0.13\textheight,width=\textwidth]{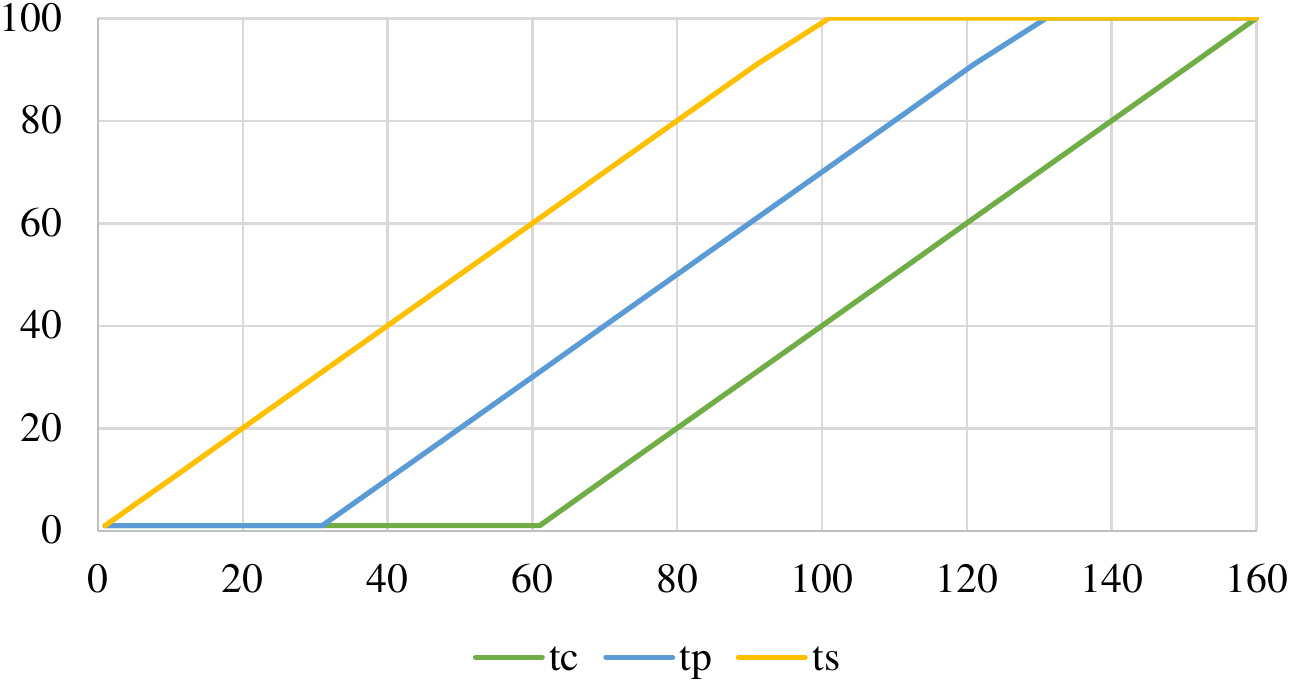}
   \caption{Partial-decoupled strategy}
   \label{fig:Partial decoupled strategy} 
\end{subfigure}
\begin{subfigure}[b]{0.3\textwidth}
   \includegraphics[height=0.13\textheight,width=\textwidth]{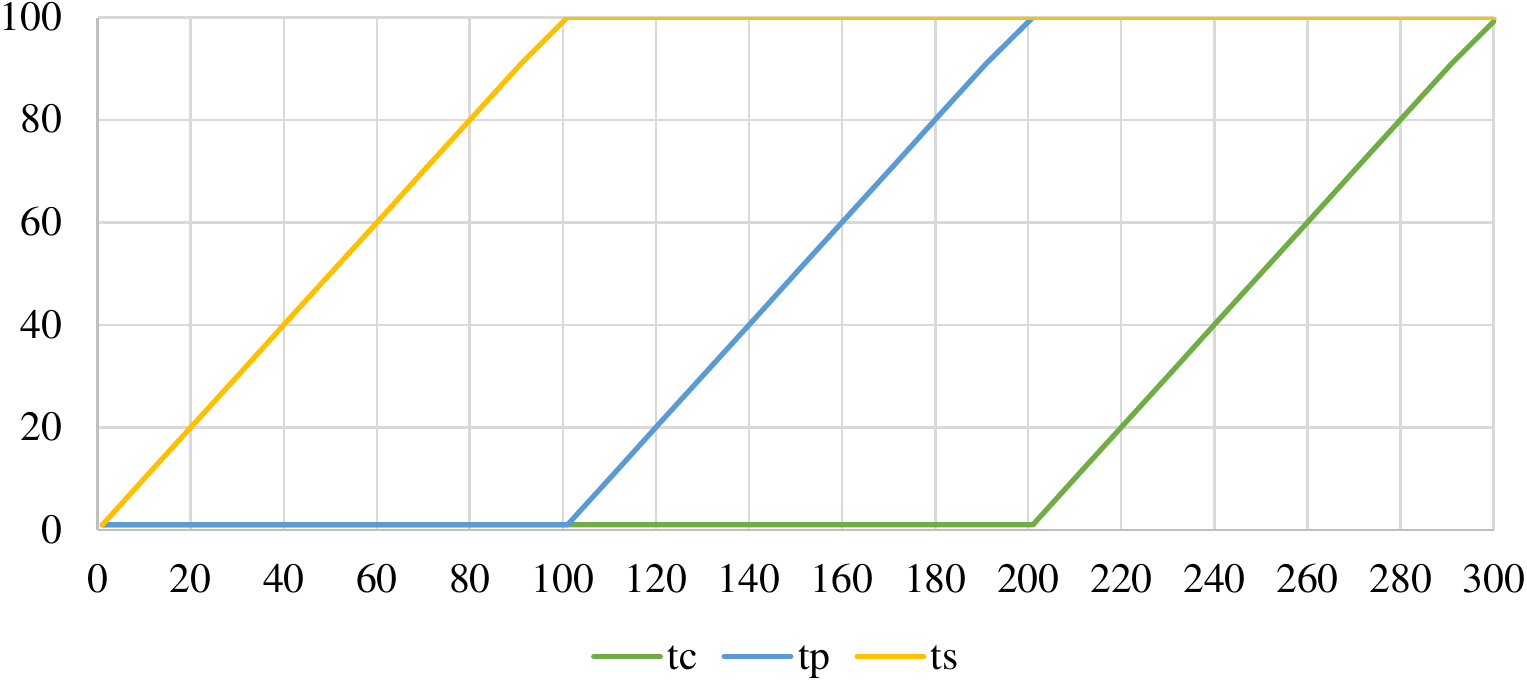}
   \caption{Sequential-decoupled strategy}
   \label{fig:Sequential-decoupled strategy} 
\end{subfigure}

\begin{subfigure}[b]{0.9\textwidth}
   \includegraphics[width=\textwidth]{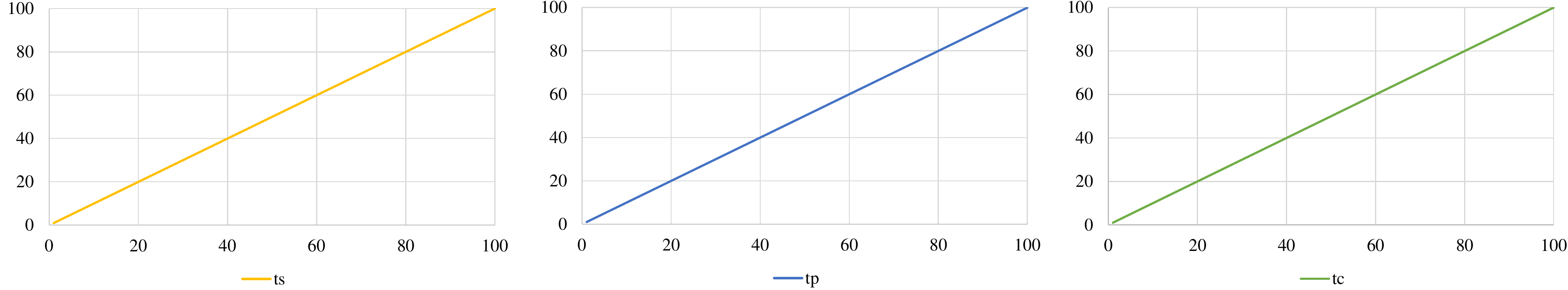}
   \caption{Parallel-decoupled strategy (ours)}
   \label{fig:Ours} 
\end{subfigure}
\caption{Illustrations of different decoupled corruption strategies. The x-axis represents the overall timestep $t$ from $1$ to $R$, while the y-axis represents individual timesteps $t_c$, $t_s$, and $t_p$ for different attributes respectively in the corresponding algorithms.}
\label{fig:Illustrations4diffusion}
\end{figure*}

In this section, we introduce more implementation details of the four different diffusion strategies in the ablation study for our proposed decoupled diffusion strategy (\ieno, the first experiment in Section 5.4 of our main paper). We provide their corresponding illustrations in Figure \ref{fig:Illustrations4diffusion}. The algorithm descriptions for \textit{Non-decoupled strategy}, \textit{Partial-decoupled strategy} and \textit{Sequential-decoupled strategy} are placed in Algorithm~\ref{alg:Non-decoupled}, Algorithm~\ref{alg:Partial decoupled} and Algorithm~\ref{alg:Sequential-decoupled}, respectively. The algorithm description for our proposed strategy, \ieno, \textit{Parallel-decoupled strategy (ours)}, has been placed in Algorithm 1 of our main paper.


\begin{algorithm}[!t]
    \caption{Non-decoupled strategy}
    \label{alg:Non-decoupled}
    \begin{algorithmic}[1]
      \Require{Max diffusion steps $T$}
      \State $l \gets $ sample a layout from the training set
      \State {$\hat{l} \gets \text{RandSelect}(l)$} \Comment{Select attributes for diffusion.}
      \State {$\hat{l} \gets [C, P, S]$} \Comment{Group $\hat{l}$ upon the semantics.}
      \State {sample $t \sim \text{Uniform}(\{1, \cdots, T\})$} 
       \For {$x$ in $C$}
       \State{$t_c \gets t$}
        \State $x = x_{t_c} \gets  \text{sample from}~q(x_{t_c}|x_0)$ 
       \EndFor
       \For {$x$ in $P$}
       \State{$t_p \gets t$}
        \State $x = x_{t_p} \gets  \text{sample from}~q(x_{t_p}|x_0)$ 
      \EndFor
       \For {$x$ in $S$}
       \State{$t_s \gets t$}
        \State $x = x_{t_s} \gets  \text{sample from}~q(x_{t_s}|x_0)$ 
       \EndFor
    \end{algorithmic}
\end{algorithm}
\begin{algorithm}[!t]
    \caption{Partial decoupled strategy}
    \label{alg:Partial decoupled}
    \begin{algorithmic}[1]
      \Require{Max diffusion steps $T$, overlap $0.3T$}
      \State $l \gets $ sample a layout from the training set
      \State {$\hat{l} \gets \text{RandSelect}(l)$} \Comment{Select attributes for diffusion.}
        \State {$\hat{l} \gets [C, P, S]$} \Comment{Group $\hat{l}$ upon the semantics.}
        \State {sample $t \sim \text{Uniform}(\{1, \cdots, 1.6T\})$} 
       \For {$x$ in $C$}
       \If{$t < 0.6T$}
       \State{$t_c \gets 1$}
       \Else
       \State{$t_c \gets t-0.6T $}
       \EndIf
       \State $x = x_{t_c} \gets  \text{sample from}~q(x_{t_c}|x_0)$ 
       \EndFor
       \For {$x$ in $P$}
       \If{$t < 0.3T$}
       \State{$t_p \gets 1$}
       \ElsIf{$t > 1.3T$}
       \State{$t_p \gets T $}
       \Else
       \State{$t_p \gets t- 0.3T$}
       \EndIf 
        \State $x = x_{t_p} \gets  \text{sample from}~q(x_{t_p}|x_0)$   
      \EndFor
       \For {$x$ in $S$}
       \If{$t < T$}
       \State{$t_s \gets t$}
       \Else
       \State{$t_s \gets T $}
       \EndIf
       \State $x = x_{t_s} \gets  \text{sample from}~q(x_{t_s}|x_0)$ 
       \EndFor
    \end{algorithmic}
\end{algorithm}
\begin{algorithm}[!t]
    \caption{Sequential-decoupled strategy}
    \label{alg:Sequential-decoupled}
    \begin{algorithmic}[1]
      \Require{Max diffusion steps $T$}
      \State $l \gets $ sample a layout from the training set
      \State {$\hat{l} \gets \text{RandSelect}(l)$} \Comment{Select attributes for diffusion.}
        \State {$\hat{l} \gets [C, P, S]$} \Comment{Group $\hat{l}$ upon the semantics.}
        \State {sample $t \sim \text{Uniform}(\{1, \cdots, 3T\})$} 
       \For {$x$ in $C$}
       \If{$t < 2T$}
       \State{$t_c \gets 1$}
       \Else
       \State{$t_c \gets t-2T $}
       \EndIf
       \State $x = x_{t_c} \gets  \text{sample from}~q(x_{t_c}|x_0)$ 
       \EndFor
       \For {$x$ in $P$}
       \If{$t < T$}
       \State{$t_p \gets 1$}
       \ElsIf{$T<t < 2T+1$}
       \State{$t_p \gets t-T $}
       \Else
       \State{$t_p \gets T$}
       \EndIf 
        \State $x = x_{t_p} \gets  \text{sample from}~q(x_{t_p}|x_0)$   
      \EndFor
       \For {$x$ in $S$}
       \If{$t < T$}
       \State{$t_s \gets t$}
       \Else
       \State{$t_s \gets T $}
       \EndIf
       \State $x = x_{t_s} \gets  \text{sample from}~q(x_{t_s}|x_0)$ 
       \EndFor
    \end{algorithmic}
\end{algorithm}

\section{More Experiment Results}
\label{sec:more_exp}

We further investigate the effectiveness of our proposed decoupled diffusion strategy by comparing different adopted noise types and decouple degrees/levels for diffusion.
All experiments are performed on the Rico dataset and the Gen-PCM task.

\subsection{Ablation Study for Noise Types}
\label{subsec:ablation_noise}

In our proposed \ours{}, we add different types of noises for the diffusion processes of different types of attributes as introduced in our main paper. Specifically, we adopt uniform noise for category $c$, while adopting Gaussian noise for location $(x, y)$ and size $(w, h)$.
In fact, it does not make sense to adopt other types of noises for category $c$ since it is hard to measure the ``distance'' between different categories. Therefore, here, we validate the effectiveness of adopting Gaussian noise for position $(x, y)$ and size $(w, h)$ by comparing it to adopting another two types of noises to diffusion. One is using the uniform noise that is the same as the one adapted to category. The other is adopting band-diagonal noise~\cite{austin2021structured}, with the one for $h$ as an example, which can be formulated as:

\begin{align}
[\alpha_t^h]_{ij} &= 1 - {\textstyle \sum_{j=0, j\neq i}^{K^h} [Q_t^h]_{ij}},\label{eq:band_diagonal_alpha} \\
[\beta_t^h]_{ij} &= \begin{cases}
\frac{1}{K} \sigma_t^h \quad &\text{if} \quad  0 < \vert i-j\vert \leq v \\
0 \quad &\text{else}\\
\end{cases}
 \label{eq:uniform_band_diagonal}
\end{align}
where $Q_t^h$ denotes the transition matrix for height $h$. $K^h$ is the number of possible values for $h$. 
$ \gamma_t^h$ and $\sigma_t^h$ are two hyperparameters that increase linearly as time $t$ increases.

\begin{table}[!t]
    \caption{Experiment results on the Rico dataset by varying the noise type on input tokens.}
    \label{tab: Corruption Strategy}
    \small
    \centering
    \resizebox{\linewidth}{!}{
    \begin{tabular}{lcccc}
      \toprule
      \multicolumn{1}{c}{Model}  &  MaxIoU  $\uparrow$        & FID  $\downarrow$             & Align.  $\downarrow$       & Overlap    $\downarrow$ \\
      \midrule
      Uniform  & 0.16 & 61.24 & 0.53 & 106.04  \\ 
      Band-diagonal  &  0.29 & 47.74 & 0.49 & 88.30  \\
      Gaussian (Ours)  & \textbf{0.59}	&\textbf{21.59}&	\textbf{0.40}	&\textbf{54.77} \\ 
      \bottomrule
    \end{tabular}
    }
\end{table}
The comparison results are in Table~\ref{eq:uniform_band_diagonal}. We can find that adopting Gaussian noise for position $(x, y)$ and size $(w, h)$ as we propose in \ours{} is the most effective design choice. This is because adopting Gaussian noise corresponds to a distance-aware noise adding (diffusion) strategy, while adopting uniform noise can not achieve this and adopting band-diagonal noise provides limited distance awareness.

\subsection{Ablation Study for Decoupling Levels}
\label{subsec:ablation_decoupling}
In our proposed diffusion strategy of \ours{}, we decouple the diffusion processes for different attributes according to attribute types.
To validate the effectiveness of this proposed design, we compare it with three other design choices with different decoupling levels: \textit{(i)}~\textit{Non-decoupling}: all attributes are corrupted with a shared timeline without decoupling. \textit{(ii)}~\textit{Element-level decoupling}: attribute tokens corresponding to the same layout element are corrupted with a shared timeline. \textit{(iii)}~\textit{Token-level decoupling}: all attribute tokens are corrupted with their individual timelines (\ieno, maximum decoupling). \textit{(iv)}~\textit{Ours}: attribute tokens of the same type are corrupted with a shared timeline as we proposed in our main paper.

\begin{table}[!t]
    \caption{Comparison results for the models with different decoupling levels on the Rico dataset.}
    \label{tab: Different level}
    \small
    \centering
    \resizebox{\linewidth}{!}{
    \begin{tabular}{lcccc}
      \toprule
      \multicolumn{1}{c}{Model}  &  MaxIoU  $\uparrow$        & FID  $\downarrow$             & Align.  $\downarrow$       & Overlap    $\downarrow$ \\
      \midrule
      Non-decoupling  & 0.56 & 29.24 & 0.43 & 60.04  \\ 
      Element-level  & 0.56 & 27.71 & 0.48 & 64.04  \\ 
      
      Token-level  & 0.58 & 23.58 & 0.41 & 57.09  \\ 
      Ours  & \textbf{0.59}	&\textbf{21.59}&	\textbf{0.40}	&\textbf{54.77} \\ 
      \bottomrule
    \end{tabular}
    }
\end{table}
The corresponding comparison results are in Table~\ref{tab: Different level}. We find that our proposed design in \ours{} is the most effective one compared to the other three. This is because we decouple the diffusion processes upon the attribute types, yielding a semantics-aware decoupling strategy.

\subsection{More Quantitative Results}

The quantitative results in Section 5.2 of our main paper are the mean values averaged over 5 runs with different random seeds. In this section, we supplement these quantitative results by further reporting their corresponding standard deviations. The results on Magazine, Rico and PinLayNet datasets are in Tables~\ref{Tab:quantitative_results_magazine}, \ref{Tab:quantitative_results_rico}, and \ref{Tab:quantitative_results_pubnet}, respectively.
\label{subsec:more_quantitative}
\begin{table*}
    \centering
    \caption{Experiment results of different layout generation subtasks on the Magazine dataset. \textit{Align.} denotes the alignment metric.}
    \renewcommand{\arraystretch}{1.2}
    \begin{small}
        \resizebox{0.7\textwidth}{!}{
            \begin{tabular}{llcccccccccccc}
                \toprule
                   \multirow{2}{*}{Subtasks}                             &    \multirow{2}{*}{Methods}                   & \multicolumn{4}{c}{Magazine}                                                                                         &                                                                                                                                                                                                  \\ \cmidrule(l){3-6}                     
                                & & MaxIoU  $\uparrow$        & FID  $\downarrow$             & Align.  $\downarrow$       & Overlap    $\downarrow$  \\  \hline
        \multirow{4}{*}{UGen} & LayoutTransformer~\cite{gupta2021layouttransformer}  & 0.18{ $\pm$ 0.03} & 47.84{ $\pm$ 1.03} & 0.59{ $\pm$ 0.03} & 47.98{ $\pm$ 0.87}  \\ 
        ~ & BLT~\cite{kong2021blt}     & 0.20{ $\pm$ 0.04} & 44.91{ $\pm$ 1.56} & 0.55{ $\pm$ 0.05} & 55.56{ $\pm$ 1.05}  \\ 
        ~ & UniLayout~\cite{jiang2022unilayout}    & 0.31{ $\pm$ 0.01} & 36.61{ $\pm$ 1.23} &0.49{ $\pm$ 0.03} & \textbf{44.50{ $\pm$ 1.02}}  \\ 
        ~ & {\ours{} (Ours)}  & \textbf{0.38{ $\pm$ 0.00}} &\textbf{32.73{ $\pm$ 0.62}} & \textbf{0.47{ $\pm$ 0.01}} & 46.43{ $\pm$ 0.98}   \\ \hline
       \multirow{4}{*}{Gen-T} & LayoutGAN++~\cite{kikuchi2021constrained}    & 0.26{ $\pm$ 0.01} & 36.35{ $\pm$ 0.77} & 0.54{ $\pm$ 0.02} & 58.44{ $\pm$ 0.73}    \\ 
        ~ & BLT~\cite{kong2021blt}   & 0.22{ $\pm$ 0.04} & 48.26{ $\pm$ 1.96} & 0.69{ $\pm$ 0.03} & 64.01{ $\pm$ 1.43}   \\ 
        ~ & UniLayout~\cite{jiang2022unilayout}    & 0.32{ $\pm$ 0.01} & 28.37{ $\pm$ 1.26} & 0.51{ $\pm$ 0.03} & 53.56{ $\pm$ 1.02} \\ 
        ~ & {\ours{} (Ours)}  & \textbf{0.36{ $\pm$ 0.01}} &\textbf{24.67{ $\pm$ 0.43}} & \textbf{0.45{ $\pm$ 0.03}} & \textbf{45.11{ $\pm$ 0.79}}   \\  \hline
        \multirow{4}{*}{Gen-TS} & BLT~\cite{kong2021blt}    & 0.33{ $\pm$ 0.03} & 22.72{ $\pm$ 1.54} & 0.59{ $\pm$ 0.01} & 61.94{ $\pm$ 1.00}    \\  
        ~ & UniLayout~\cite{jiang2022unilayout}    & 0.35{ $\pm$ 0.01} & 19.35{ $\pm$ 0.72} & 0.58{ $\pm$ 0.03} & 56.43{ $\pm$ 1.02}  \\  
        ~ & {\ours{} (Ours)}  & \textbf{0.37{ $\pm$ 0.02}} & \textbf{17.65{ $\pm$ 0.57}} & \textbf{0.45{ $\pm$ 0.02}} & \textbf{44.25{ $\pm$ 0.56}}   \\   \hline
        \multirow{3}{*}{Gen-TR} & CLG-LO~\cite{kikuchi2021constrained}     & 0.27{ $\pm$ 0.01} & 33.88{ $\pm$ 1.31} & 0.59{ $\pm$ 0.05} & 59.43{ $\pm$ 1.39}  \\  
        ~ & UniLayout~\cite{jiang2022unilayout}     & 0.36{ $\pm$ 0.01} & \textbf{19.24{ $\pm$ 0.88}} & 0.54{ $\pm$ 0.03} & 49.61{ $\pm$ 0.74}   \\  
        ~ & {\ours{} (Ours)}  & \textbf{0.39{ $\pm$ 0.02} }& 20.58{ $\pm$ 0.45} & \textbf{0.48{ $\pm$ 0.03}} & \textbf{47.27{ $\pm$ 0.80}}  \\  \hline
        \multirow{3}{*}{Refinement} & RUITE & 0.24{ $\pm$ 0.03} & 44.27{ $\pm$ 1.32} & 0.64{ $\pm$ 0.05} & 54.26{ $\pm$ 1.41}  \\  
        ~ & UniLayout~\cite{jiang2022unilayout}   & 0.33{ $\pm$ 0.01} & 19.78{ $\pm$ 0.70} & 0.49{ $\pm$ 0.03} & 49.02{ $\pm$ 0.96}  \\  
        ~ & {\ours{} (Ours)}  & \textbf{0.39{ $\pm$ 0.00}} & \textbf{14.95{ $\pm$ 0.68}} & \textbf{0.42{ $\pm$ 0.01} }& \textbf{37.22{ $\pm$ 0.51}}   \\   \hline
        \multirow{3}{*}{Completion} & LayoutTransformer~\cite{gupta2021layouttransformer}  & 0.17{ $\pm$ 0.03} & 39.36{ $\pm$ 1.83} & 0.67{ $\pm$ 0.02} & 55.32{ $\pm$ 0.76}  \\  
        ~ & UniLayout~\cite{jiang2022unilayout}    &0.23{ $\pm$ 0.01} & 28.78{ $\pm$ 1.60} & 0.52{ $\pm$ 0.03} & 46.43{ $\pm$ 0.64}  \\  
        ~ & {\ours{} (Ours)}  & \textbf{0.38{ $\pm$ 0.00}} & \textbf{24.35{ $\pm$ 0.43}} & \textbf{0.49{ $\pm$ 0.01}} & \textbf{39.26{ $\pm$ 0.49}}  \\  \hline
                        Gen-PM      &  \multirow{4}{*}{\ours{} (Ours)}  & 0.38{ $\pm$ 0.01} & 27.33{ $\pm$ 0.54} & 0.47{ $\pm$ 0.01}& 39.02{ $\pm$ 0.39}  \\
                Gen-CM      &           & 0.37{ $\pm$ 0.02} & 28.74{ $\pm$ 0.79} & 0.51{ $\pm$ 0.01}& 43.25{ $\pm$ 0.71} \\
                Gen-PC      &            & 0.37{ $\pm$ 0.01} & 22.56{ $\pm$ 0.62} & 0.47{ $\pm$ 0.00}& 42.95{ $\pm$ 0.30}  \\
                Gen-PCM      &            & 0.37{ $\pm$ 0.01} & 24.45{ $\pm$ 1.21} & 0.49{ $\pm$ 0.01}& 44.41{ $\pm$ 0.89} \\\hline
                GT         &-  & 0.41 & 9.89 & 0.43 & 34.27    \\                     
                \specialrule{1.1pt}{1pt}{0pt}
            \end{tabular}}                     
    \end{small}
    \label{Tab:quantitative_results_magazine}
\end{table*}

\begin{table*}
    \centering
    \caption{Experiment results of different layout generation subtasks on the Rico dataset. \textit{Align.} denotes the alignment metric.}
    \renewcommand{\arraystretch}{1.2}
    \begin{small}
        \resizebox{0.7\textwidth}{!}{
            \begin{tabular}{llcccccccccccc}
                \toprule
                   \multirow{2}{*}{Subtasks}                             &    \multirow{2}{*}{Methods}   & \multicolumn{4}{c}{Rico}                                                                                                                                                                                                                \\ \cmidrule(l){3-6}          
                                & & MaxIoU  $\uparrow$        & FID  $\downarrow$             & Align.  $\downarrow$       & Overlap    $\downarrow$ \\  \hline
        \multirow{4}{*}{UGen} & LayoutTransformer~\cite{gupta2021layouttransformer}   & 0.46{ $\pm$ 0.03}  & 46.64{ $\pm$ 0.97}  & 0.66{ $\pm$ 0.02}  & 64.10{ $\pm$ 0.94}  \\ 
        ~ & BLT~\cite{kong2021blt}     & 0.51{ $\pm$ 0.01}  & 33.81{ $\pm$ 1.56}  & 0.59{ $\pm$ 0.04}  & 67.33{ $\pm$ 0.71}   \\ 
        ~ & UniLayout~\cite{jiang2022unilayout}    & \textbf{0.62{ $\pm$ 0.01}}  & 26.68{ $\pm$ 0.74}  & 0.40{ $\pm$ 0.03}  & 59.26{ $\pm$ 0.76}    \\ 
        ~ & {\ours{} (Ours)} & \textbf{0.62{ $\pm$ 0.01}}  & \textbf{26.06{ $\pm$ 0.40}}  & \textbf{0.36{ $\pm$ 0.03}}  & \textbf{56.35{ $\pm$ 0.71}}   \\ \hline
       \multirow{4}{*}{Gen-T} & LayoutGAN++~\cite{kikuchi2021constrained}  & 0.46{ $\pm$ 0.00}  & 34.43{ $\pm$ 1.13}  & 0.58{ $\pm$ 0.02}  & 59.85{ $\pm$ 0.85}   \\ 
        ~ & BLT~\cite{kong2021blt}   &0.44{ $\pm$ 0.03}  & 39.64{ $\pm$ 1.71}  & 0.57{ $\pm$ 0.01}  & 56.83{ $\pm$ 1.45}  \\ 
        ~ & UniLayout~\cite{jiang2022unilayout}   &0.55{ $\pm$ 0.01}  & 18.06{ $\pm$ 0.70}  & 0.48{ $\pm$ 0.03}  & 57.92{ $\pm$ 0.94}    \\ 
        ~ & {\ours{} (Ours)}  & \textbf{0.58{ $\pm$ 0.01}}  & \textbf{16.64{ $\pm$ 0.46}}  & \textbf{0.39{ $\pm$ 0.03}}  & \textbf{55.87{ $\pm$ 0.64}}     \\  \hline
        \multirow{4}{*}{Gen-TS} & BLT~\cite{kong2021blt}      & 0.51{ $\pm$ 0.03}  & 42.88{ $\pm$ 1.26}  & 0.46{ $\pm$ 0.02}  & 57.74{ $\pm$ 0.47}   \\  
        ~ & UniLayout~\cite{jiang2022unilayout}   & 0.55{ $\pm$ 0.01}  &20.42{ $\pm$ 0.40}  & 0.49{ $\pm$ 0.02}  & 58.72{ $\pm$ 0.36}   \\  
        ~ & {\ours{} (Ours)}  &\textbf{ 0.62{ $\pm$ 0.01}}  & \textbf{12.59{ $\pm$ 0.37}}  & \textbf{0.35{ $\pm$ 0.01}}  & \textbf{55.92{ $\pm$ 0.39}}   \\   \hline
        \multirow{3}{*}{Gen-TR} & CLG-LO~\cite{kikuchi2021constrained}     & 0.38{ $\pm$ 0.02}  & 38.89{ $\pm$ 0.79}  &0.54{ $\pm$ 0.02}  & \textbf{56.51{ $\pm$ 0.80}}    \\  
        ~ & UniLayout~\cite{jiang2022unilayout}     & 0.57{ $\pm$ 0.01}  & 26.38{ $\pm$ 0.92}  &0.46{ $\pm$ 0.02}  & 66.93{ $\pm$ 0.52}   \\  
        ~ & {\ours{} (Ours)}   & \textbf{0.61{ $\pm$ 0.01}}  & \textbf{16.98{ $\pm$ 0.40}}  & \textbf{0.39{ $\pm$ 0.01}}  & 58.75{ $\pm$ 0.66}    \\  \hline
        \multirow{3}{*}{Refinement} & RUITE & 0.46{ $\pm$ 0.03}  & 36.70{ $\pm$ 0.70}  &0.57{ $\pm$ 0.02}  & 64.13{ $\pm$ 1.94}   \\  
        ~ & UniLayout~\cite{jiang2022unilayout}   & 0.56{ $\pm$ 0.01}  & 24.41{ $\pm$ 0.57}  & 0.42{ $\pm$ 0.01}  & 56.04{ $\pm$ 0.65}    \\  
        ~ & {\ours{} (Ours)}  & \textbf{0.62{ $\pm$ 0.00}}  & \textbf{13.19{ $\pm$ 0.40}}  & \textbf{0.33{ $\pm$ 0.01}}  & \textbf{52.17{ $\pm$ 0.58}}    \\   \hline
        \multirow{3}{*}{Completion} & LayoutTransformer~\cite{gupta2021layouttransformer}  & 0.46{ $\pm$ 0.03}  & 36.15{ $\pm$ 0.64}  & 0.66{ $\pm$ 0.02}  & 67.10{ $\pm$ 0.65}    \\  
        ~ & UniLayout~\cite{jiang2022unilayout}    & 0.59{ $\pm$ 0.01}  & 25.18{ $\pm$ 0.42}  & 0.45{ $\pm$ 0.04}  & 55.99{ $\pm$ 0.92}   \\  
        ~ & {\ours{} (Ours)} & \textbf{0.60{ $\pm$ 0.01}}  &\textbf{ 16.42{ $\pm$ 0.66}}  & \textbf{0.36{ $\pm$ 0.03}}  & \textbf{53.15{ $\pm$ 0.64}}    \\  \hline
                        Gen-PM      &  \multirow{4}{*}{\ours{} (Ours)}  & 0.58{ $\pm$ 0.01}  & 21.64{ $\pm$ 0.50}  & 0.38{ $\pm$ 0.03}  & 56.56{ $\pm$ 0.85}   \\
                Gen-CM      &            & 0.57{ $\pm$ 0.03}  & 26.15{ $\pm$ 0.46}  & 0.38{ $\pm$ 0.03}  & 57.74{ $\pm$ 0.79}      \\
                Gen-PC      &             & 0.60{ $\pm$ 0.01}  & 18.13{ $\pm$ 0.33}  & 0.36{ $\pm$ 0.00}  & 53.67{ $\pm$ 0.46}      \\
                Gen-PCM      &            & 0.59{ $\pm$ 0.01}  & 21.59{ $\pm$ 0.75}  & 0.40{ $\pm$ 0.01}  & 54.77{ $\pm$ 0.70}     \\\hline
                GT         &-   & 0.66 & 7.05 & 0.26 & 49.86      \\                     
                \specialrule{1.1pt}{1pt}{0pt}
            \end{tabular}}                     
    \end{small}
    \label{Tab:quantitative_results_rico}
\end{table*}

\begin{table*}
    \centering
    \caption{Experimental results of different layout generation subtasks on the PublayNet dataset. \textit{Align.} denotes the alignment metric.}
    \renewcommand{\arraystretch}{1.2}
    \begin{small}
        \resizebox{0.7\textwidth}{!}{
            \begin{tabular}{llcccccccccccc}
                \toprule
                   \multirow{2}{*}{Subtasks}     &    \multirow{2}{*}{Methods} & \multicolumn{4}{c}{PubLayNet}                                                                                                                         \\ \cmidrule(l){3-6}      
                                & & MaxIoU  $\uparrow$        & FID  $\downarrow$             & Align.  $\downarrow$       & Overlap    $\downarrow$\\  \hline
        \multirow{4}{*}{UGen} & LayoutTransformer~\cite{gupta2021layouttransformer} &  0.32{ $\pm$ 0.01} & 49.72{ $\pm$ 1.81} & 0.37{ $\pm$ 0.01} & 36.63{ $\pm$ 0.43} \\ 
        ~ & BLT~\cite{kong2021blt}  &   0.34{ $\pm$ 0.01} & 48.24{ $\pm$ 0.68} & 0.27{ $\pm$ 0.04} & 42.79{ $\pm$ 0.53} \\ 
        ~ & UniLayout~\cite{jiang2022unilayout}  &   0.33{ $\pm$ 0.01} & 32.29{ $\pm$ 0.56} & \textbf{0.22{ $\pm$ 0.02}} & 22.19{ $\pm$ 0.22} \\ 
        ~ & {\ours{} (Ours)} & \textbf{0.46{ $\pm$ 0.01}} & \textbf{25.94{ $\pm$ 0.41}} & 0.25{ $\pm$ 0.01} & \textbf{19.83{ $\pm$ 0.19}}  \\ \hline
       \multirow{4}{*}{Gen-T} & LayoutGAN++~\cite{kikuchi2021constrained} &  0.36{ $\pm$ 0.01} & 30.48{ $\pm$ 0.75} &0.19{ $\pm$ 0.02} & 32.80{ $\pm$ 0.42}  \\ 
        ~ & BLT~\cite{kong2021blt} &  0.37{ $\pm$ 0.03} & 44.86{ $\pm$ 0.66} & 0.21{ $\pm$ 0.03} & 38.21{ $\pm$ 0.43} \\ 
        ~ & UniLayout~\cite{jiang2022unilayout} &  0.41{ $\pm$ 0.01} & 27.34{ $\pm$ 0.96} & 0.20{ $\pm$ 0.02} & 20.98{ $\pm$ 0.37} \\ 
        ~ & {\ours{} (Ours)} & \textbf{0.44{ $\pm$ 0.01}} & \textbf{20.69{ $\pm$ 0.36}} &\textbf{0.15{ $\pm$ 0.02}} & \textbf{16.88{ $\pm$ 0.39}}  \\  \hline
        \multirow{4}{*}{Gen-TS} & BLT~\cite{kong2021blt}  &  0.40{ $\pm$ 0.00} & 24.32{ $\pm$ 0.73} & 0.16{ $\pm$ 0.02} & 31.06{ $\pm$ 0.58}  \\  
        ~ & UniLayout~\cite{jiang2022unilayout}  &   0.43{ $\pm$ 0.01} & 27.47{ $\pm$ 0.85} & \textbf{0.16{ $\pm$ 0.01}} & 23.82{ $\pm$ 0.66}  \\  
        ~ & {\ours{} (Ours)} &  \textbf{0.47{ $\pm$ 0.01}} & \textbf{19.02{ $\pm$ 0.35}} & \textbf{0.16{ $\pm$ 0.01}} & \textbf{10.09{ $\pm$ 0.38}} \\   \hline
        \multirow{3}{*}{Gen-TR} & CLG-LO~\cite{kikuchi2021constrained}  &    0.38{ $\pm$ 0.02} & 31.87{ $\pm$ 0.82} & 0.21{ $\pm$ 0.03} & 34.39{ $\pm$ 0.59} \\  
        ~ & UniLayout~\cite{jiang2022unilayout}  &   \textbf{0.46{ $\pm$ 0.01}} & 27.73{ $\pm$ 0.84} & 0.17{ $\pm$ 0.02} & 27.35{ $\pm$ 0.61}  \\  
        ~ & {\ours{} (Ours)}& 0.44{ $\pm$ 0.01} & \textbf{19.54{ $\pm$ 0.43}} & \textbf{0.16{ $\pm$ 0.01}} & \textbf{21.28{ $\pm$ 0.54}} \\  \hline
        \multirow{3}{*}{Refinement} & RUITE& 0.32{ $\pm$ 0.01} & 41.72{ $\pm$ 0.99} & 0.49{ $\pm$ 0.01} & 35.74{ $\pm$ 1.89}  \\  
        ~ & UniLayout~\cite{jiang2022unilayout}  & 0.44{ $\pm$ 0.01} & 22.34{ $\pm$ 0.87} & 0.11{ $\pm$ 0.01} & 27.23{ $\pm$ 0.73}  \\  
        ~ & {\ours{} (Ours)}   & \textbf{0.48{ $\pm$ 0.01}} & \textbf{15.28{ $\pm$ 0.59}} & \textbf{0.10{ $\pm$ 0.01}} & \textbf{13.05{ $\pm$ 0.42}}  \\   \hline
        \multirow{3}{*}{Completion} & LayoutTransformer~\cite{gupta2021layouttransformer}  & 0.32{ $\pm$ 0.01} & 41.72{ $\pm$ 0.81} & 0.37{ $\pm$ 0.01} & 39.81{ $\pm$ 0.20}  \\  
        ~ & UniLayout~\cite{jiang2022unilayout}     & 0.41{ $\pm$ 0.01} & 32.04{ $\pm$ 0.55} & 0.19{ $\pm$ 0.02} & 22.90{ $\pm$ 0.29}  \\  
        ~ & {\ours{} (Ours)}   & \textbf{0.44{ $\pm$ 0.01}} & \textbf{25.31{ $\pm$ 0.60}} & \textbf{0.10{ $\pm$ 0.00}} & \textbf{19.45{ $\pm$ 0.28}}  \\  \hline
                        Gen-PM      &  \multirow{4}{*}{\ours{} (Ours)}  & 0.46{ $\pm$ 0.01} & 23.58{ $\pm$ 0.18} & 0.10{ $\pm$ 0.00}& 14.11{ $\pm$ 0.29} \\
                Gen-CM      &            & 0.44{ $\pm$ 0.01} & 24.94{ $\pm$ 0.24} & 0.11{ $\pm$ 0.01} & 16.26{ $\pm$ 0.47} \\
                Gen-PC      &           & 0.50{ $\pm$ 0.01} & 16.42{ $\pm$ 0.18} & 0.09{ $\pm$ 0.01} & 12.51{ $\pm$ 0.11} \\
                Gen-PCM      &            & 0.42{ $\pm$ 0.00} & 25.76{ $\pm$ 0.59} & 0.14{ $\pm$ 0.01} & 19.68{ $\pm$ 0.58} \\\hline
                GT         &-    & 0.64 & 9.38 & 0.008 & 5.18      \\                     
                \specialrule{1.1pt}{1pt}{0pt}
            \end{tabular}}                     
    \end{small}
    \label{Tab:quantitative_results_pubnet}
\end{table*}

\subsection{More Qualitative Results}
\label{subsec:more_qualitative}

\begin{figure*}[!t]
    \centering
    \includegraphics[width=\linewidth]{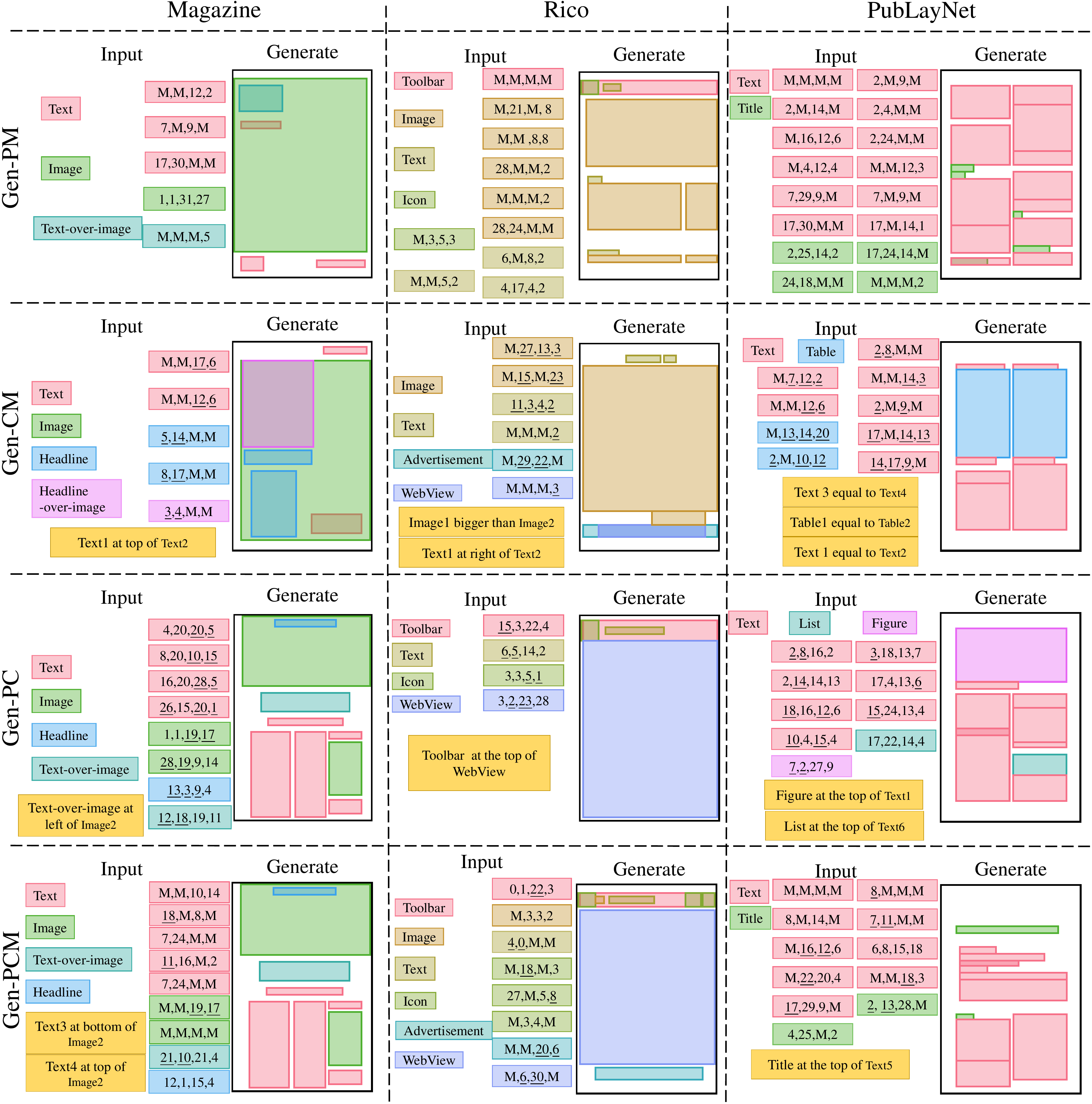}
    \caption{Qualitative results on general layout generation tasks(Gen-CM,Gen-PC,Gen-PC,Gen-PCM).M represents missing attribute, \underline{x} represents coarse attribute, and x represents precise attribute. } 
    \label{fig:suppQ}
\end{figure*}

In our main paper, we introduce four more general layout generation task settings which can cover the existing ones defined in previous works. They are \textit{Gen-PM}, \textit{Gen-CM}, \textit{Gen-PC} and \textit{Gen-PCM}. All six available layout generation subtasks defined in previous could be all considered as the special cases of them. In this section, we provide the qualitative generation results on \textit{Gen-PM},  \textit{Gen-CM}, \textit{Gen-PC} and \textit{Gen-PCM} in Figure~\ref{fig:suppQ}. Note that to the best of our knowledge, no previous works can support them so that we are not allowed to compare our proposed \ours{} with others on these newly proposed task settings. We make the first endeavour to achieve such comprehensive versatility.  
\subsection{Results of Rendered Images}
\begin{figure*}[!t]
    \centering
    \includegraphics[width=\linewidth]{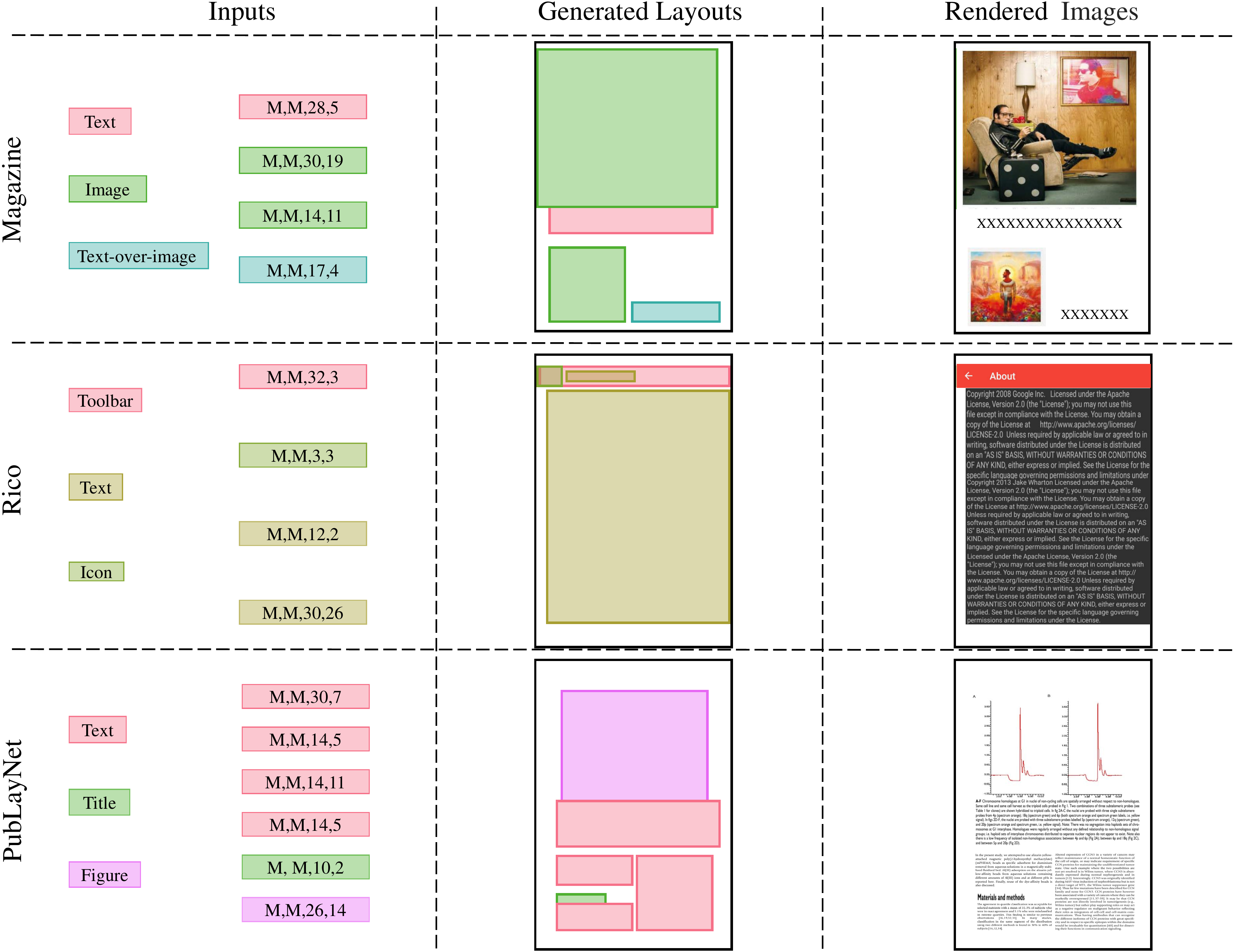}
    \caption{Rendered Images. The generated layouts are from three models trained on Magazine, Rico, and PubLayNet, respectively.} 
    \label{fig:suppR}
\end{figure*}

We show the results of rendered images upon the generated layouts by our proposed \ours{} in Figure~\ref{fig:suppR}. These rendered images help visually demonstrate the quality of our generated layouts.
\label{subsec:rendered_images}

\end{document}